\documentclass{article}

\usepackage{microtype}
\usepackage{graphicx}
\usepackage{subcaption}
\usepackage{booktabs} 

\usepackage{hyperref}
\hypersetup{breaklinks=true}
\usepackage{tcolorbox}

\usepackage{amsmath}
\usepackage{amssymb}
\usepackage{mathtools}
\usepackage{amsthm}
\usepackage{enumitem}

\usepackage{multirow}
\usepackage{makecell}

\usepackage[capitalize,noabbrev]{cleveref}

\usepackage[preprint]{icml2026}

\theoremstyle{plain}

\theoremstyle{definition}

\theoremstyle{remark}

\usepackage[textsize=tiny]{todonotes}

\icmltitlerunning{ASTER: Agentic Scaling with Tool-integrated Extended Reasoning}

\begin{document}

\twocolumn[
  \icmltitle{ASTER: Agentic Scaling with Tool-integrated Extended Reasoning}



  \icmlsetsymbol{equal}{*}
  
  \begin{icmlauthorlist}
    \icmlauthor{Xuqin Zhang}{equal,huawei,KeyLab,NJUAI}
    \icmlauthor{Quan He}{equal,huawei}
    \icmlauthor{Zhenrui Zheng}{cuhk}
    \icmlauthor{Zongzhang Zhang}{KeyLab,NJUAI}
    \icmlauthor{Xu He}{huawei}
    \icmlauthor{Dong Li}{huawei}
    
  \end{icmlauthorlist}

  \icmlaffiliation{KeyLab}{National Key Laboratory for Novel Software Technology, Nanjing University, China}
  \icmlaffiliation{NJUAI}{School of Artificial Intelligence, Nanjing University, China}
  \icmlaffiliation{huawei}{Department of Foundation Model, 2012 Labs, Huawei}
  \icmlaffiliation{cuhk}{The Chinese University of Hong Kong, Shenzhen}

  \icmlcorrespondingauthor{Quan He}{jiaqibbq@gmail.com}
  \icmlcorrespondingauthor{Zongzhang Zhang}{zzzhang@nju.edu.cn}


  \icmlkeywords{Reinforcement Learning, Tool-Integrated Reasoning, Large Language Models, Agentic AI, Mathematical Reasoning, Cold-start Strategy, Behavioral Prior}

  \vskip 0.3in
]



\printAffiliationsAndNotice{}  



\begin{abstract}
Reinforcement learning (RL) has emerged as a dominant paradigm for eliciting long-horizon reasoning in Large Language Models (LLMs). 
However, scaling Tool-Integrated Reasoning (TIR) via RL remains challenging due to \emph{interaction collapse}: a pathological state where models fail to sustain multi-turn tool usage, instead degenerating into heavy internal reasoning with only trivial, post-hoc code verification. 
We systematically study three questions: (i) how cold-start SFT induces an agentic, tool-using behavioral prior, (ii) how the interaction density of cold-start trajectories shapes exploration and downstream RL outcomes, and (iii) how the RL interaction budget affects learning dynamics and generalization under varying inference-time budgets. 
We then introduce \textbf{ASTER}  (\textbf{A}gentic \textbf{S}caling with \textbf{T}ool-integrated \textbf{E}xtended \textbf{R}easoning), a framework that circumvents this collapse through a targeted cold-start strategy prioritizing interaction-dense trajectories. 
We find that a small expert cold-start set of just 4K interaction-dense trajectories yields the strongest downstream performance, establishing a robust prior that enables superior exploration during extended RL training. 
Extensive evaluations demonstrate that ASTER-4B achieves state-of-the-art results on competitive mathematical benchmarks, reaching 90.0\% on AIME 2025, surpassing leading frontier open-source models, including DeepSeek-V3.2-Exp.

\end{abstract}

\section{Introduction}

\begin{figure*}[!t]  \centering
  \includegraphics[width=0.9\textwidth]{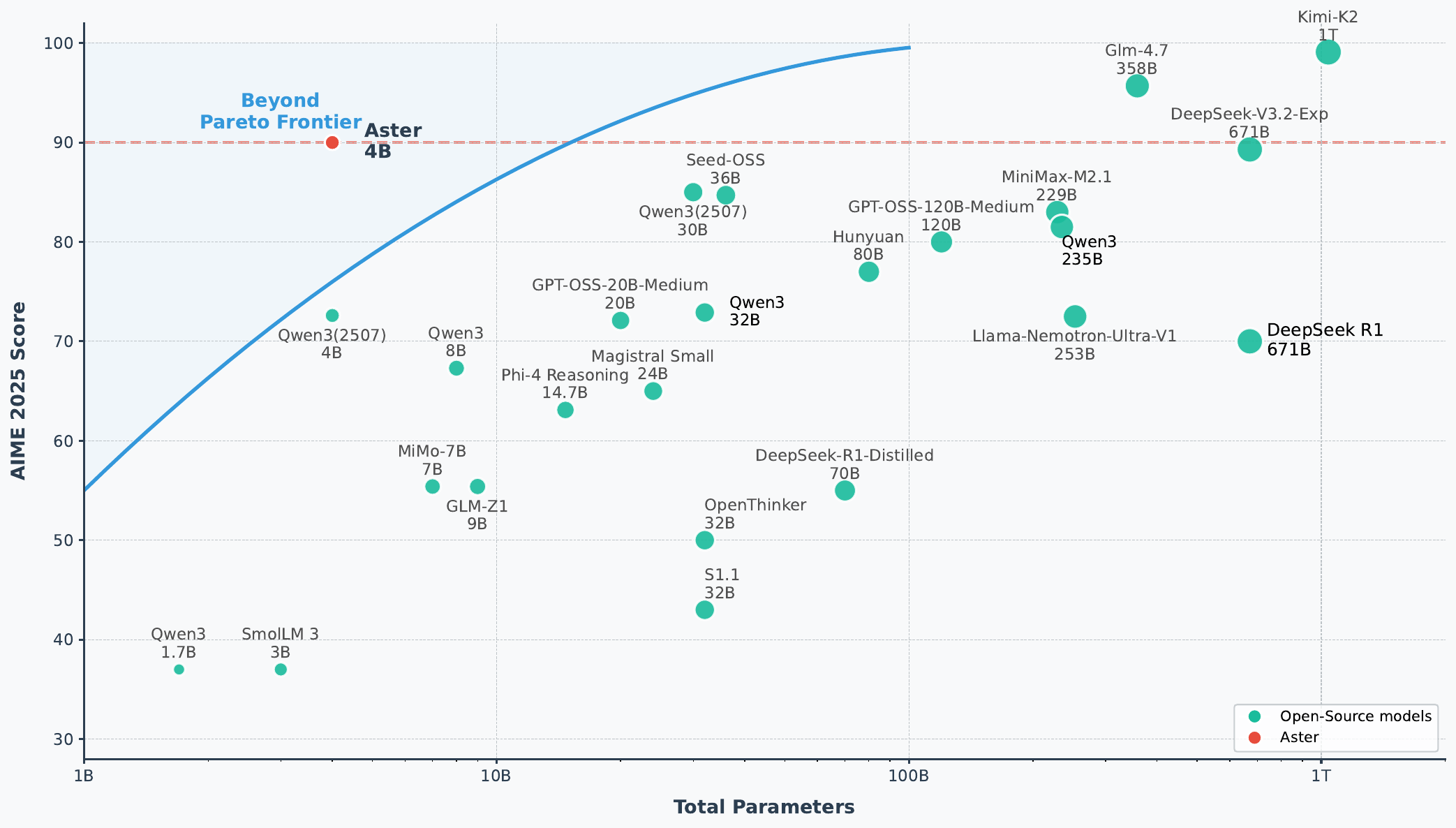}
  \caption{ASTER demonstrates remarkable efficiency, surpassing much larger and stronger models on the challenging
  AIME 2025 benchmark. It achieves a score of 90.0, outperforming DeepSeek-V3.2-exp (89.3/671B).}
  \vspace{-15pt}
\end{figure*}

Large Language Models (LLMs) have achieved significant reasoning improvements through Reinforcement Learning (RL), which facilitates the refinement of long chains of thought (CoT)~\citep{wei_chainofthought,yao_treethoughts}.
Recent systems such as OpenAI o1~\cite{openai_openaio1card} and DeepSeek R1~\cite{deepseekai_deepseekr1} demonstrate the effectiveness of this approach.
By employing analytical strategies such as error correction, problem decomposition, and iterative refinement, these models exhibit strong capabilities in complex reasoning tasks~\citep{yeo_demystifyinglongchainofthoughtreasoning}.

While long CoT has achieved success, text-only reasoning remains inherently fragile, as the absence of external verification allows minor errors to cascade beyond the reach of self-correction mechanisms. 
TIR overcomes this by augmenting language models with executable tools, 
leveraging external feedback to enable exact computation and intermediate verification, thereby ensuring reliability in long-horizon reasoning tasks \cite{lin_understandingtoolintegratedreasoning}.

Existing approaches to TIR primarily differ in how tool-use behavior is initialized.
Zero Tool-Integrated Reasoning (ZeroTIR) seeks to elicit tool usage directly from pretrained base models via RL, without relying on supervised cold-start data.
Despite its conceptual simplicity, ZeroTIR suffers from two fundamental limitations.
From the perspective of optimization, multi-turn TIR training is frequently unstable. Tool execution introduces non-stationary feedback and distributional drift toward low-probability tokens, which compounds across turns and often results in gradient explosion and poor convergence behavior \cite{wang_ragen, mai_agentrlscalinglaw, xue_simpletir}.
From the perspective of capability, RL predominantly amplifies pre-existing competencies rather than instilling fundamentally new ones \cite{zhang_interplaypretrainingmidtrainingrl}. As a consequence, ZeroTIR remains bounded by the representational and procedural limits of the underlying base model, and consistently struggles with long-horizon planning, coordinated tool invocation, and program-level error correction.

An alternative paradigm initializes TIR via Supervised Fine-Tuning (SFT) on carefully chosen, tool-augmented trajectories before RL~\citep{feng_retool,shang_rstar2agent,yu_demystifyingreinforcementlearningagentic}. 
These trajectories are commonly obtained either by rewriting text-only CoT traces into code-interleaved~\cite{feng_retool} or by synthesizing tool-using demonstrations with code-specialized teacher models \cite{yu_demystifyingreinforcementlearningagentic}. 
Despite their empirical success, existing cold-start designs introduce strong inductive biases through trajectory construction. ReTool~\cite{feng_retool} largely preserve the original text-only reasoning structure, limiting the emergence of genuinely tool-driven planning. 
In contrast, trajectories produced by code-specialized teachers~\cite{yu_demystifyingreinforcementlearningagentic} often feature short, reactive reasoning. This contradicts the deliberate, System-2 behavior expected in slow-thinking models.
These inductive biases often lead to \emph{interaction collapse} during subsequent RL training, a pathological failure mode in which models fail to sustain multi-turn tool usage and instead degenerate into extensive internal reasoning followed by trivial, post-hoc code verification.

In this work, we introduce \textbf{ASTER} (\textbf{A}gentic \textbf{S}caling with \textbf{T}ool-integrated \textbf{E}xtended \textbf{R}easoning), a two-stage framework that combines a targeted cold-start supervised fine-tuning phase with RL to elicit long-horizon, multi-turn tool-using behavior.
Rather than optimizing for immediate post-SFT accuracy, we study how cold-start SFT induces a behavioral prior for multi-turn tool use. Specifically, we ask:
\begin{itemize}
  \item \textbf{RQ1 (Cold-start priors).} What role does cold-start SFT play in inducing an agentic, tool-using behavioral prior, and how do different cold-start designs shape downstream tool-use strategies and performance under RL?
  \item \textbf{RQ2 (Interaction density).} How does the interaction density of cold-start trajectories influence exploration behavior and subsequent RL outcomes?
  \item \textbf{RQ3 (RL interaction budget).} How does the RL interaction budget impact learning dynamics and generalization, especially when inference-time budgets vary?
\end{itemize}

Through extensive experiments, we identify \emph{interaction density} as the key property of the cold-start prior. 
Concretely, we synthesize tool-augmented trajectories with GPT-OSS-20B and then curate a small expert dataset (4K) with more than nine tool-interaction turns as the cold-start SFT prior.
By initializing the model with long, iterative tool-use trajectories, ASTER induces a high-entropy behavioral prior, thereby preserving exploration capacity during subsequent RL. 
While this design choice yields lower post-SFT accuracy compared to reasoning-centric initialization, it crucially prevents premature convergence to short-horizon, suboptimal policies. 
As a result, the model is able to more fully exploit capability scaling during RL, ultimately surpassing baselines trained on sparse, short-horizon trajectories.

Building on this interaction-dense prior, ASTER further employs a multi-stage RL procedure to stabilize and refine extended reasoning loops. 
Empirically, ASTER-4B achieves state-of-the-art results, reaching 85.0\% accuracy on AIME 2025 and 73.3\% on HMMT 2025 under 30K inference budget. 
Despite its compact size, it consistently outperforms the 235B-parameter Qwen3-235B-A22B-Thinking across a diverse suite of competitive mathematics benchmarks. 
Moreover, ASTER exhibits strong test-time scaling: with increased inference budgets up to 90K, performance improves to 90.0\% on AIME 2025, confirming that interaction-dense training unlocks scalable agentic intelligence.

\section{Preliminaries}
\label{sec:preliminary}

\begin{table*}[!t]
  \centering
  \scalebox{0.95}{
  \begin{tabular}{lcccc}
  \toprule
  \textbf{Model} & 
  \textbf{AIME2024} & 
  \textbf{AIME2025} & 
  \textbf{HMMT2025} & 
  \textbf{BeyondAIME} \\
  & \textcolor{gray}{avg@16} & \textcolor{gray}{avg@16} & \textcolor{gray}{avg@16} & \textcolor{gray}{avg@16} \\
  \midrule
  \multicolumn{5}{l}{\textit{Text-Only Reasoning Models}} \\
  Qwen3-1.7B-Thinking & 47.5 & 38.3 & 25.8 & 20.8 \\
  Qwen3-4B-Thinking-2507 & 76.7 & 72.7 & 48.1 & 43.6 \\
  OpenAI o3-mini (medium) & 79.6 & 77.0 & 53.0 & -- \\
  POLARIS-4B-Preview & 81.2 & 79.4 & 58.7 & -- \\
  OpenReasoning-Nemotron-7B & 84.7 & 78.2 & 63.5 & -- \\
  Qwen3-235B-A22B-Thinking & \underline{85.7} & 81.5 & 62.5 & -- \\
  \midrule
  \multicolumn{5}{l}{\textit{Agentic Reasoning Models}} \\
  ReTool-32B & 72.5 & 54.3 & -- & -- \\
  rStar2-Agent-14B & 80.6 & 69.8 & 52.7 & -- \\
  DemyAgent-4B & 72.6 & 70.0 & 52.9$^\dagger$ & 35.3$^\dagger$ \\
  \midrule
  ASTER-1.7B-SFT & 19.4 & 19.0 & 11.3 & 6.4 \\
  ASTER-1.7B & 64.6 & 59.6 & 47.5 & 26.3 \\
  ASTER-4B-SFT & 62.5 & 54.6 & 43.3 & 27.4 \\
  ASTER-4B & 82.3 & \underline{85.0} & \underline{73.3} & \underline{53.9} \\
  \quad \textit{w/ 90K Inference Budget} & \textbf{85.8} & \textbf{90.0} & \textbf{77.1} & \textbf{61.7} \\
  \bottomrule
\end{tabular}
  }
  \caption{Performance comparison on competitive mathematical benchmarks. We report average $\text{avg}@16$ results. 
  All evaluation metrics adhere to the DeepSeek-R1 assessment framework (temperature=$0.6$, $\text{top}\_\text{p}=0.95$). We prioritize official performance data where available; results marked with ${}^\dagger$ denote our independent evaluation using officially recommended configurations \textbf{restricted to a $30\text{K}$ inference budget}. The best scores are highlighted in \textbf{bold}, and the second-best are \underline{underlined}.}
  \vspace{-15pt}
  \label{tab:transposed_benchmarks}
\end{table*}

\paragraph{Supervised Fine-Tuning}

SFT adapts pre-trained LLMs to specialized tasks via next-token prediction on high-quality instruction-response pairs. 
In the context of agentic reasoning, SFT often functions as a cold-start phase to initialize multi-turn tool-use capabilities.
Common strategies for constructing such training data involve rewriting text-only CoT solutions into code-interleaved formats~\citep{feng_retool} or leveraging code-specialized teacher models to synthesize tool-using demonstrations~\citep{yu_demystifyingreinforcementlearningagentic}. 
This initialization establishes the necessary formats and baseline interaction capabilities for subsequent reinforcement learning.

\paragraph{Reinforcement Learning}

While SFT establishes the foundational behavioral patterns, it is limited by the quality and diversity of the demonstration data.
RL is subsequently employed to generalize beyond imitation, optimizing the model's reasoning capabilities through exploration and feedback. In the context of mathematical reasoning and agentic tasks, RL shifts the training objective from minimizing prediction error to maximizing a reward signal—typically defined by the correctness of the final answer or the validity of intermediate execution steps. 
Group Relative Policy Optimization (GRPO)~\cite{shao_deepseekmath-grpo} has emerged as a prominent algorithm. Unlike traditional methods like PPO that rely on computationally expensive value function critics, GRPO optimizes the policy by sampling a group of G outputs $\{o_1, o_2, ..., o_G\}$ for a single query $q$. It utilizes the relative performance within this group as a baseline, which significantly reduces memory overhead and stabilizes training.

The optimization objective maximizes the following surrogate loss:
\begin{align*}
\mathcal{J}(\theta)
&= \mathbb{E}_{q \sim \mathcal{D},\, \{o_i\}_{i=1}^G \sim \pi_{\theta_{\text{old}}}(\cdot|q)} \Bigg[
\frac{1}{\sum_{i=1}^{G} |o_i|} \sum_{i=1}^{G} \sum_{t=1}^{|o_i|} \\
&\min\!\Big(
r_{i,t}(\theta)\hat{A}_{i,t},\,
\operatorname{clip}\!\big(r_{i,t}(\theta),1-\epsilon,1+\epsilon\big)\hat{A}_{i,t}
\Big)
\Bigg],
\end{align*}
where $r_{i,t}(\theta) = \frac{\pi_{\theta}(o_{i,t}|q, o_{i,<t})}{\pi_{\theta_{\text{old}}}(o_{i,t}|q, o_{i,<t})}$ is the probability ratio between the current policy $\pi_{\theta}$ and the old policy $\pi_{\theta_{\text{old}}}$ (i.e., the policy used for sampling before the update), and $\hat{A}_{i,t}$ is the advantage computed based on the relative rewards within the group.
\section{Towards Best Agentic Scaling Recipe}

In this section, we examine how post-training pipelines, specifically cold-start SFT design and RL interaction budgets, shape the emergence of agentic reasoning in LLMs.
We study whether cold-start SFT induces a behavioral prior that governs how models plan, explore, and interact with tools during downstream RL.
We organize our empirical analysis around the following research questions:

\begin{itemize}[leftmargin=*]

  \item \textbf{RQ1 (Role of cold-start SFT and behavioral priors).} 
  How do cold-start SFT design choices (e.g., trajectory patterns) shape the induced behavioral prior, and how does this prior influence tool-use strategies and performance in subsequent RL?    
  \item \textbf{RQ2 (Effect of interaction density of SFT).} 
  How does the interaction density of cold-start SFT affect the induced behavioral prior and the performance in subsequent RL?
  \item \textbf{RQ3 (Effect of interaction budget of RL training).} 
  How does the interaction budget during RL training affect learning dynamics and final performance, and how does it impact test-time performance when the inference-time tool budget is varied?
\end{itemize}

\subsection{Experimental Setup}

\paragraph{Training data}\label{sec:training_data}
We construct our SFT dataset from two primary sources: (i) Skywork-OR1-RL-Data from Skywork-OR1\footnote{\href{https://huggingface.co/datasets/Skywork/Skywork-OR1}{Skywork-OR1 dataset}}~\citep{he_skyworkopenreasoner1},
and (ii) 93K problems from the Art of Problem Solving (AoPS) forums collected by OpenMathReasoning~\citep{moshkov_aimo2winningsolutionbuilding}.
We retain only English problems whose ground-truth final answers are positive integers.
We then generate tool-augmented solutions using GPT-OSS-20B-high and retain only instances with both a correct final answer and a 100\% tool execution success rate, yielding an initial dataset of 45K trajectories. 
For RL training, we employ DAPO-Math-17K\footnote{\href{https://huggingface.co/datasets/BytedTsinghua-SIA/DAPO-Math-17k}{DAPO-Math-17K dataset}}, which contains 17K prompts paired with integer ground-truth answers.

\paragraph{Base Models}
We use Qwen3-1.7B-Thinking and Qwen3-4B-Thinking-2507 as our base models.

\paragraph{Cold-start SFT}
We employ \texttt{LLaMA-Factory}\footnote{\href{https://github.com/hiyouga/LlamaFactory}{GitHub: hiyouga/LlamaFactory}} for multi-turn instruction tuning.
Training is conducted for 6 epochs with a global batch size of 128 and a maximum context length of 32K tokens.
We adopt a constant learning rate: $3 \times 10^{-5}$ for Qwen3-1.7B/4B.

\paragraph{Reinforcement Learning}
Our framework inherits VeRL\footnote{\href{https://github.com/volcengine/verl}{GitHub: volcengine/verl}} as a submodule, ensuring compatibility with upstream updates.
We perform RL training using GRPO. 
Following \cite{yu_dapo}, we optimize only the token-level policy gradient objective with Clip-Higher ($\epsilon_h = 0.28$) and do not include a KL divergence penalty.
For data collection, we use a rollout prompt batch size of 128 and sample $G=8$ trajectories per prompt to form the group baseline.
We employ gradient accumulation with a mini-batch size of 64, yielding exactly two parameter updates per rollout step.  Unless otherwise specified, step refers to a \textit{rollout step}. We use a unified conversation template throughout our experiments.
To support long-horizon tool use and iterative self-correction, we set the maximum allowed tool invocations to 50 per trajectory.
\paragraph{Evaluation Protocol}
We evaluate mathematical reasoning proficiency on four competitive benchmarks: AIME2024, AIME2025, HMMT2025 (February), and BeyondAIME~\citep{seed_seed15thinking}.
Unless otherwise specified, we use temperature 0.6 and $\text{top}\_\text{p}=0.95$ for all evaluations.
All other evaluation parameters follow the recommended settings from the respective baseline papers.
By default, we use a maximum context length of 32K tokens for evaluation.

\paragraph{Baselines}
We compare our models against several state-of-the-art baselines, including OpenReasoning-Nemotron-7B~\citep{ahmad_OpenReasoning-Nemotron}, Qwen3-235B-A22B-Thinking~\citep{yang_qwen3}, POLARIS-4B-Preview~\citep{an_polaris4bpreview}, ReTool-32B~\citep{feng_retool}, as well as rStar2-Agent-14B~\citep{shang_rstar2agent} and DemyAgent-4B~\citep{yu_demystifyingreinforcementlearningagentic}. 
All baseline performance numbers are reported directly from their respective publications.

\subsection{Role of cold-start SFT and behavioral priors (RQ1)}

\begin{figure*}[!t]
    \centering
    \begin{subfigure}[t]{0.48\textwidth}
      \centering
      \includegraphics[width=\textwidth]{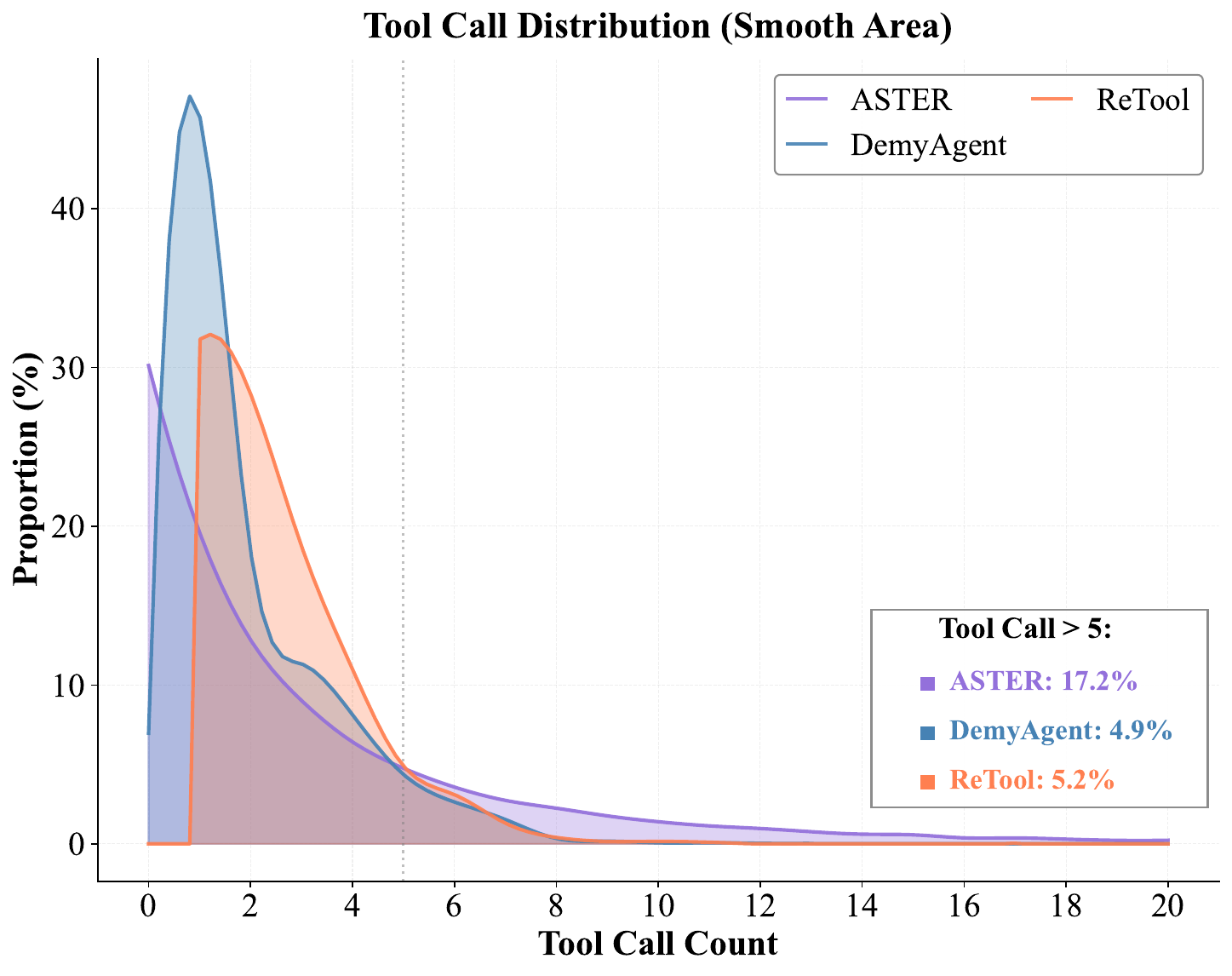}
      \caption{Tool call count distribution across different cold-start datasets.}
      \label{fig:tool_call_distribution}
    \end{subfigure}
    \hfill
    \begin{subfigure}[t]{0.48\textwidth}
      \centering
      \includegraphics[width=\textwidth]{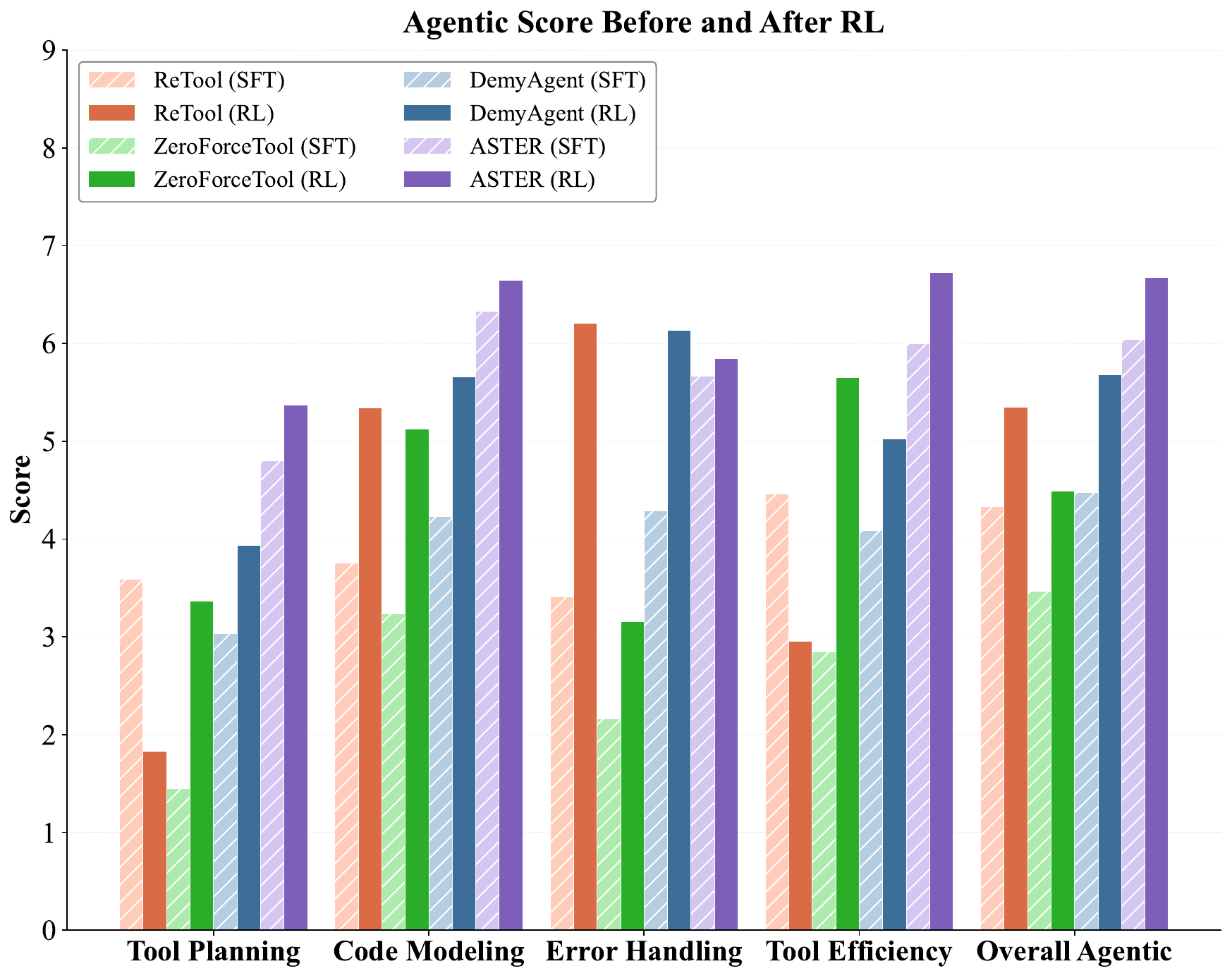}
      \caption{Agentic judge comparison.}
      \label{fig:agentic_judge_comparison}
    \end{subfigure}
    \caption{Distributional properties of cold-start datasets. (a) Tool-call count distribution reveals that ReTool and DemyAgent datasets are heavily biased toward sparse interactions (1--2 tool calls), while our dataset contains more long-horizon trajectories.
     (b) Different cold-start strategies induce distinct behavioral priors, using agentic judger to evaluate their agentic capabilities.}
    \vspace{-15pt}
\end{figure*}

To answer RQ1, we conduct a controlled comparison with Qwen3-4B-2507-Thinking as the unified base model, varying only the cold-start initialization prior to RL. We consider the following strategies:

\begin{itemize}[leftmargin=*]
    \item \textbf{Zero}: Direct RL training without cold-start SFT. The reward assigns $1$ to correct final answers and $0$ otherwise.
    \item \textbf{ZeroForceTool}: Direct RL training without cold-start SFT, but with a tool-constrained reward. The reward assigns $1$ only if the final answer is correct \emph{and} explicit tool usage is observed; otherwise, it assigns $0$.
    \item \textbf{ReTool}: Cold-start SFT on ReTool\footnote{\href{https://huggingface.co/datasets/JoeYing/ReTool-SFT}{ReTool-SFT dataset}}, which follows a ``stitch-style'' synthesis paradigm that rewrites computational steps from trajectories into code snippets.
    \item \textbf{DemyAgent}: Cold-start SFT on DemyAgent\footnote{\href{https://huggingface.co/datasets/Gen-Verse/Open-AgentRL-SFT-3K}{Open-AgentRL-SFT-3K dataset}}, consisting of end-to-end synthetic trajectories generated by Qwen3-Coder-30B-A3B.
    \item \textbf{ASTER}: Cold-start SFT on our 45K curated dataset (see \cref{sec:training_data}).
\end{itemize}

To characterize the differences among cold-start datasets, we analyze their tool call distributional properties. \Cref{fig:tool_call_distribution} reports the distribution of tool-call counts.
ReTool and DemyAgent are strongly skewed toward sparse tool use: most trajectories contain only 1--2 tool calls, and fewer than 2\% exceed 10 invocations (ReTool is as low as 0.1\%). In contrast, our SFT dataset substantially increases the proportion of long-horizon trajectories with intensive tool use.
To rule out dataset difficulty as a confounding factor, we measure per-problem accuracy with 8 samples from Qwen3-4B-2507-Thinking; \Cref{fig:difficulty_distribution_appendix} shows comparable difficulty, suggesting the key difference lies in interaction patterns rather than problem hardness.

\begin{figure*}[!t]
  \centering
  \begin{subfigure}[b]{0.48\textwidth}
    \centering
    \includegraphics[width=\textwidth]{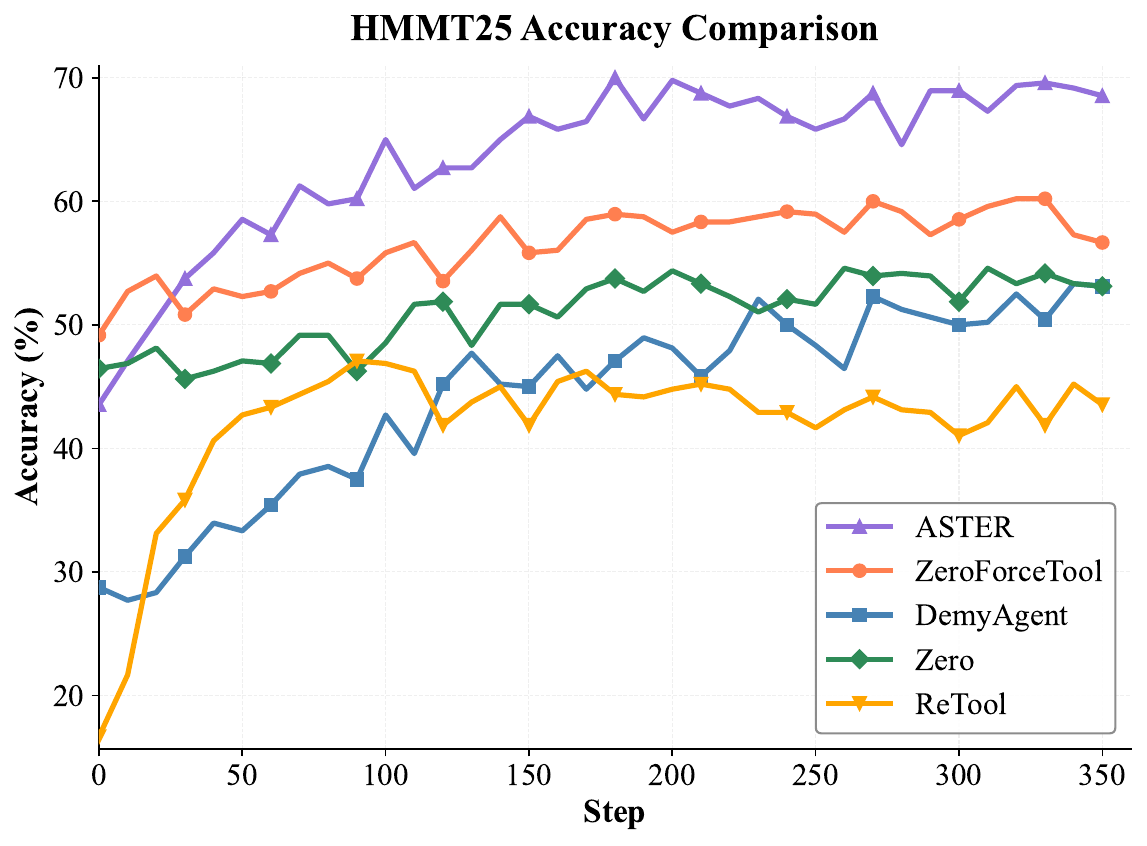}
    \caption{HMMT25 accuracy comparison.}
    \label{fig:hmmt25_accuracy}
  \end{subfigure}
  \hfill
  \begin{subfigure}[b]{0.48\textwidth}
    \centering
    \includegraphics[width=\textwidth]{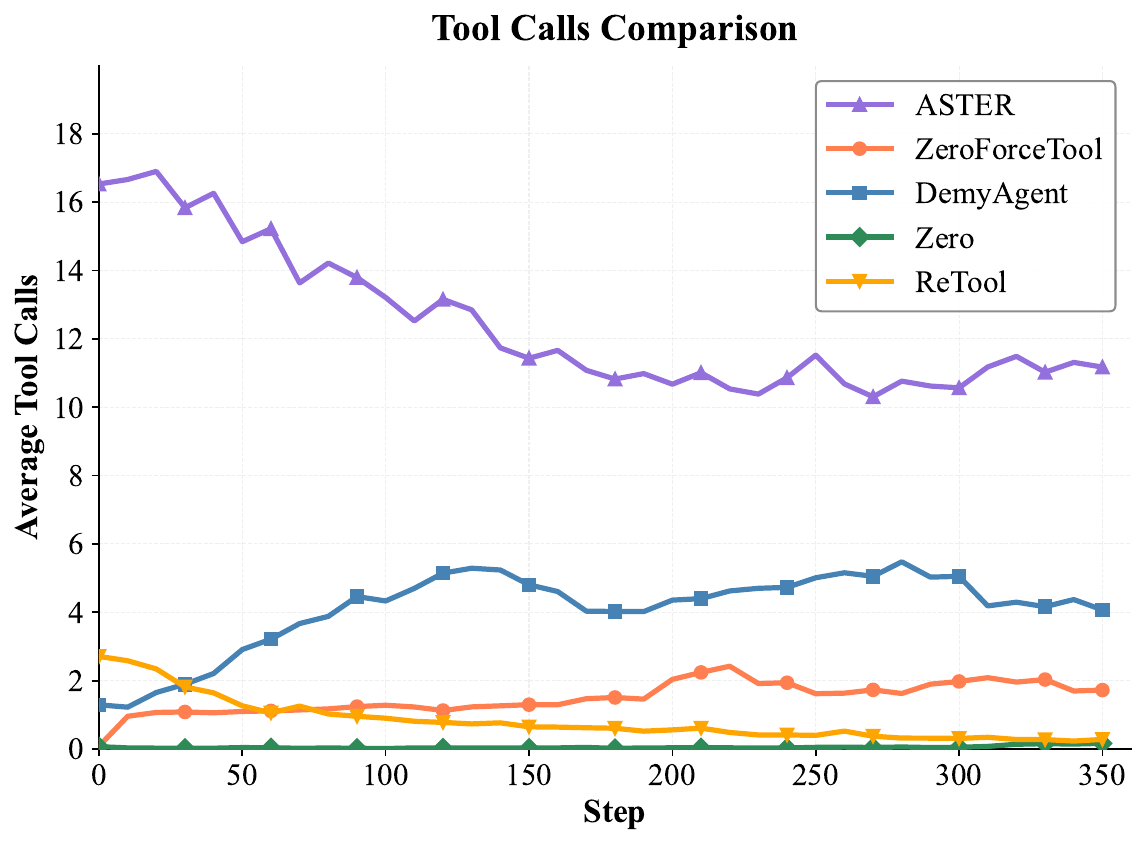}
    \caption{Tool calls comparison.}
    \label{fig:tool_calls_comparison}
  \end{subfigure}
  \caption{RQ1 analysis: Cold-start strategies shape downstream RL dynamics. (a) and (b) illustrate the impact of the synthetic teacher patterns, showing performance gaps and distinct tool-use strategies.}
  \label{fig:rq1_coldstart_dynamics}
  \vspace{-10pt}
\end{figure*}

To assess how initialization shapes intrinsic agentic behaviors after cold-start SFT and subsequent RL, we use an agentic judger based on GPT-5 with a fixed rubric and prompt template (\cref{sec:agentic_judger}) that scores \emph{planning}, \emph{code modeling}, \emph{error handling}, \emph{tool efficiency}, and \emph{overall agentic capability}. 
As shown in \cref{fig:agentic_judge_comparison}, all cold-started variants achieve higher scores than \textbf{Zero} across dimensions, and \textbf{ASTER} attains the strongest and most consistent improvements, indicating that different cold-start SFT initializations induce distinct \emph{behavior priors}.

\Cref{fig:rq1_coldstart_dynamics} summarizes the RL training dynamics in terms of task performance and tool-call numbers.
\textbf{ASTER} exhibits a distinctive pattern: it sustains intensive tool use throughout training while consistently outperforming the baselines.

In contrast, the baselines show clear limitations.
\textbf{Zero} persistently gravitates toward minimal tool usage.
\textbf{ZeroForceTool} initially increases tool usage, but often regresses to shallow verification behaviors, failing to acquire genuine multi-step interaction and exhibiting interaction collapse.
\textbf{ReTool} similarly drops to low tool usage after an initial rise and rapidly plateaus at suboptimal performance.
Finally, although \textbf{DemyAgent} reaches moderate tool usage ($\approx 4$ calls), its short-horizon training trajectories limit the learning of long interaction chains required for complex tasks, constraining final accuracy.
Overall, cold-start is not universally beneficial: when demonstrations are dominated by sparse, short-horizon tool interactions (as in ReTool and DemyAgent), the induced prior can bias learning toward shallow verification behaviors and limit downstream RL gains, failing to outperform \textbf{Zero}.


\begin{tcolorbox}[colback=black!5!white,colframe=black!50!black,title=\textbf{Takeaway 1: Cold-start SFT Determines the Behavioral Prior for RL}]
The effect of cold-start SFT is not uniform; it crucially depends on the \emph{interaction patterns} encoded in the demonstrations.
Sparse and short-horizon trajectories induce a behavioral prior that favors shallow, verification-style tool use, which constrains policy improvement under RL.
In contrast, incorporating a modest fraction of long-horizon, interaction-dense trajectories (ASTER) is sufficient to guide learning toward sustained multi-turn tool use and enable stronger downstream performance.
\end{tcolorbox}

\subsection{Effect of interaction density of SFT (RQ2)} 
\label{sec:rq2}


To study the effect of interaction density in cold-start SFT, we stratify the 45K dataset described in \cref{sec:training_data} by the number of tool invocations per trajectory.
Specifically, we construct three subsets with at most 1 tool call (22K trajectories), at most 5 tool calls (37K trajectories), and at least 9 tool calls (4K trajectories).
We cold-start the model with each subset and then run the same RL pipeline.
For this study, we use an 18K context length for both RL training and evaluation.

\begin{figure}[!t]
    \centering
    \includegraphics[width=0.48\textwidth]{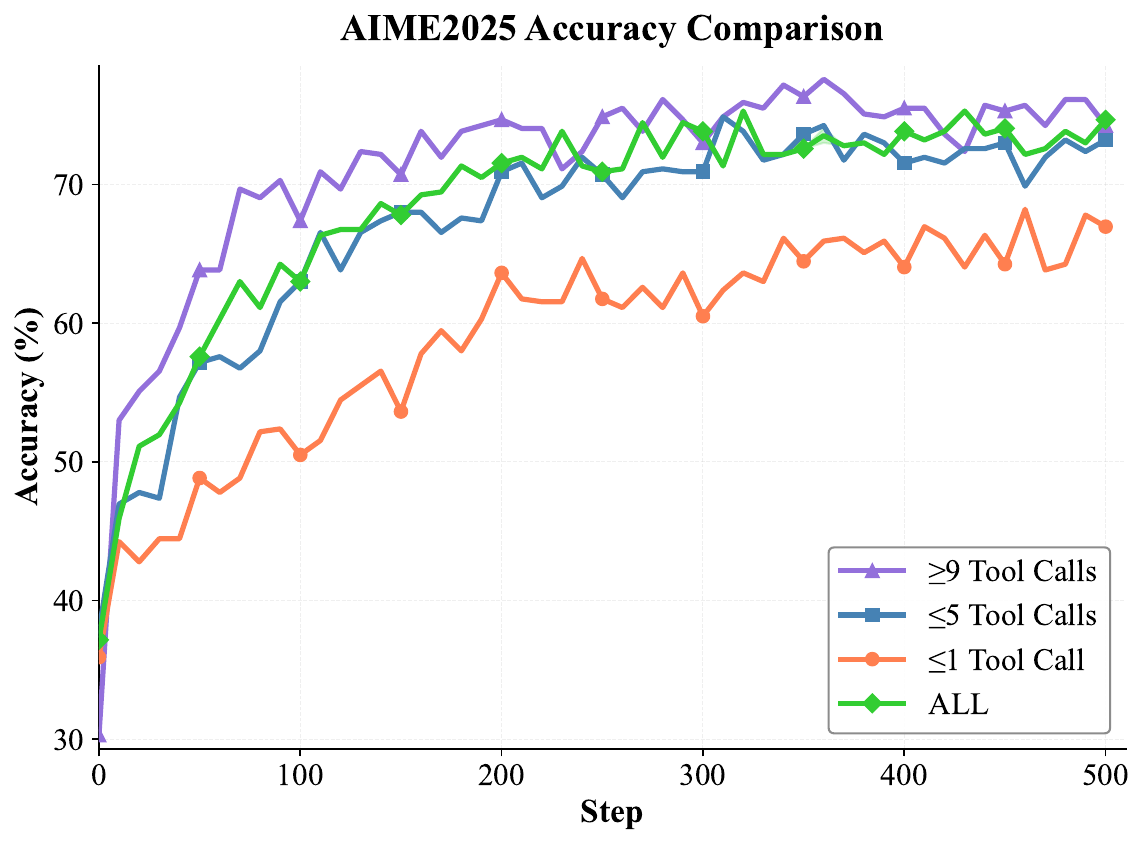}
    \caption{Tool intensive SFT achieves higher performance on AIME25.}
    \label{fig:rq2_coldstart_dynamics}
\vspace{-15pt}
\end{figure}

\Cref{fig:rq2_coldstart_dynamics} plots AIME25 accuracy during RL; \cref{sec:appendix_rq2} reports the corresponding training entropy.
We observe a consistent ordering in final accuracy: initializing from interaction-dense trajectories ($\ge 9$ tool calls) performs best, followed by the full dataset, then the $\le 5$ subset, with the $\le 1$ subset worst.
Beyond final accuracy, the $\ge 9$ initialization maintains higher entropy throughout training, indicating broader exploration rather than early collapse.
Taken together, these results suggest that tool-intensive cold-start SFT shapes the subsequent RL dynamics toward more persistent tool use, which helps the policy discover higher-quality solutions.

\begin{tcolorbox}[colback=black!5!white,colframe=black!50!black,title=\textbf{Takeaway (RQ2): Interaction Density Matters}]
    Cold-start SFT with interaction-dense trajectories establishes a stronger tool-using behavioral prior, leading to persistently higher training entropy during RL and, in turn, superior final task performance.
\end{tcolorbox}

\subsection{Effect of interaction budget of RL training (RQ3)}
\label{sec:rq3}

RQ3 examines how the RL interaction budget shapes emergent reasoning behavior and test-time scaling, comparing policies learned under constrained versus generous training-time tool budgets.

We cold-start Qwen3-4B-2507-Thinking using the 4K high-interaction trajectories introduced in \cref{sec:rq2}.
During RL training, we vary only the \textit{maximum number of tool invocations per trajectory}, setting the budget to either 10 or 50 calls, while keeping all other hyperparameters identical to the main configuration.
All RL training is conducted with an 18K context length.

At evaluation time, models are tested with an expanded 32K context window.
We systematically vary the \textit{inference-time tool budget} from 2 to 256 calls per trajectory, allowing us to characterize how training-time interaction constraints affect performance scaling across different deployment regimes.

\begin{figure}[!t]
  \centering
  \includegraphics[width=0.48\textwidth]{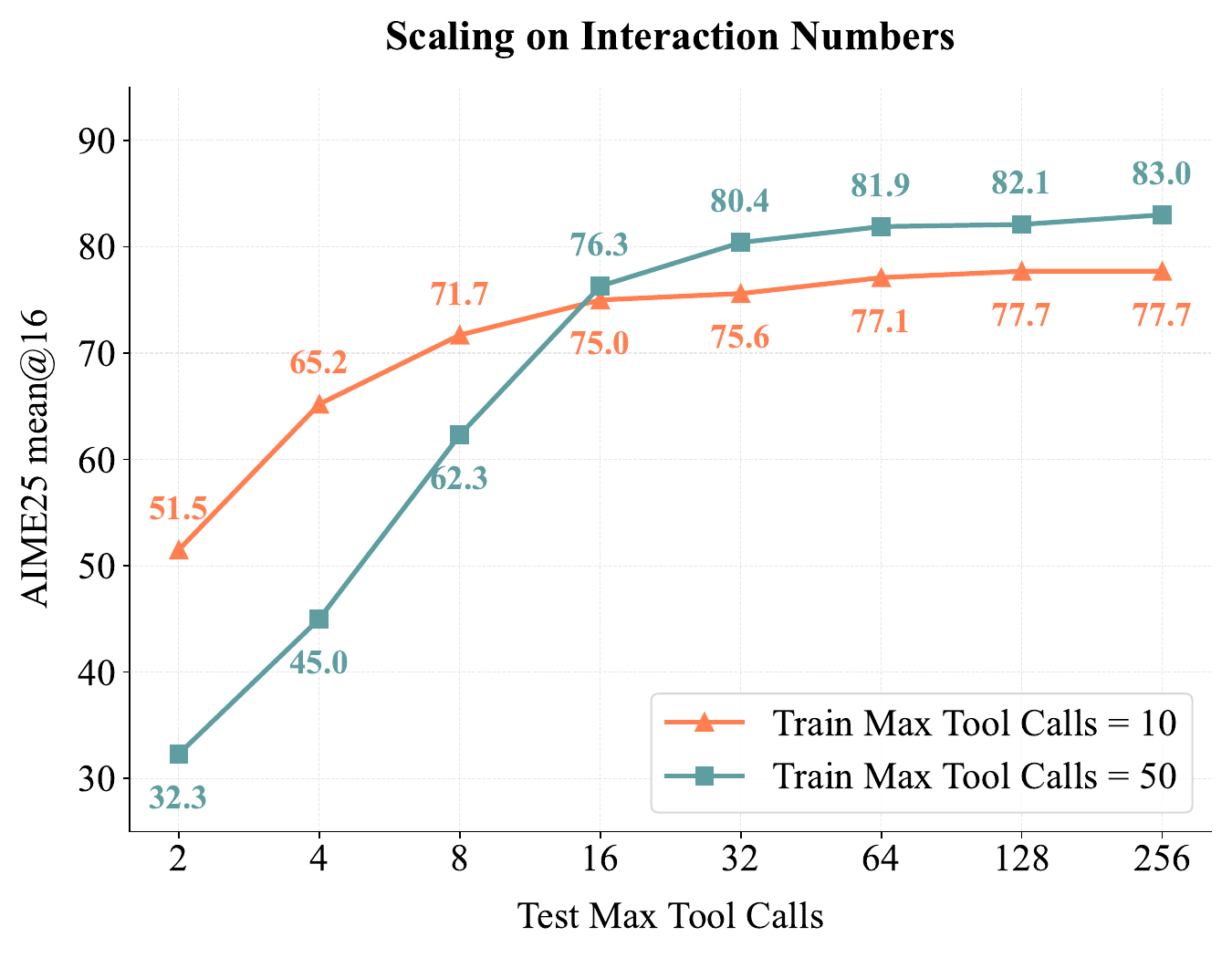}
  \caption{Test-time performance scaling as a function of inference-time tool budget for models trained under different interaction constraints. Models trained with higher interaction budgets (50 calls) dominate under large inference budgets, while constrained-training models perform better when inference is severely limited.}
  \label{fig:interaction_scaling}
  \vspace{-15pt}
\end{figure}

\Cref{fig:interaction_scaling} reveals a clear budget-alignment effect.
The scaling curves intersect at intermediate inference budgets, beyond which the relative advantage reverses.
When ample inference interaction is available, models trained with generous interaction capacity achieve substantially higher peak performance, reaching 83.0\% accuracy at 256 tool calls, compared to 77.7\% for constrained-training models.
In contrast, under tight inference budgets, models trained with limited interaction capacity are markedly more effective, achieving 51.5\% accuracy with only 2 tool calls, versus 32.3\% for models trained with higher budgets.


Overall, these results underscore the importance of aligning training-time and deployment-time interaction budgets in outcome-only RL for tool-augmented reasoning.
Mismatch between training and inference constraints can significantly degrade performance, while budget alignment enables models to operate closer to their optimal regime.

\begin{tcolorbox}[colback=black!5!white,colframe=black!50!black,title=\textbf{Takeaway (RQ3): Training-Time Interaction Budget Matters}]
    Larger training-time interaction budgets yield better test-time scaling when the inference-time tool budget increases.
\end{tcolorbox}

\section{ASTER Recipe}

Building on the insights from RQ1--RQ3, we present a concrete instantiation of ASTER.

Following RQ1 and RQ2, we train on the 4K curated subset with more than nine interaction turns.
Following RQ3, we cap the maximum number of tool calls per training trajectory at 50.

To further improve training stability and efficiency, we adopt a two-stage RL curriculum with an increasing maximum context length: 18K tokens in Stage 1 and 32K tokens in Stage 2.
Before transitioning to Stage 2, we filter out prompts that consistently produce correct outcomes across all sampled trajectories in the final epoch, 
thereby focusing subsequent training on more challenging instances.  

To evaluate the effectiveness of ASTER, we present a comparative analysis in \cref{tab:transposed_benchmarks}. The results demonstrate that ASTER achieves state-of-the-art performance across all evaluated benchmarks, significantly outperforming both text-only reasoning models and existing agentic reasoning systems. Notably, ASTER-4B with a 90K inference budget achieves 85.8\%, 90.0\%, 77.1\%, and 61.7\% on AIME2024, AIME2025, HMMT2025, and BeyondAIME, respectively.
When compared to existing agentic reasoning models, ASTER exhibits even more pronounced advantages: ASTER-4B outperforms the best agentic baseline (rStar2-Agent-14B) by 5.2\%, 20.2\%, and 24.2\% on AIME2024, AIME2025, and HMMT2025, respectively. 
Remarkably, despite being trained from a 4B model, ASTER-4B achieves performance that matches or exceeds much larger models, including Qwen3-235B-A22B-Thinking and OpenAI o3-mini, while maintaining superior tool-integrated reasoning capabilities.

\begin{figure*}[!t]  \centering
  \includegraphics[width=1\textwidth]{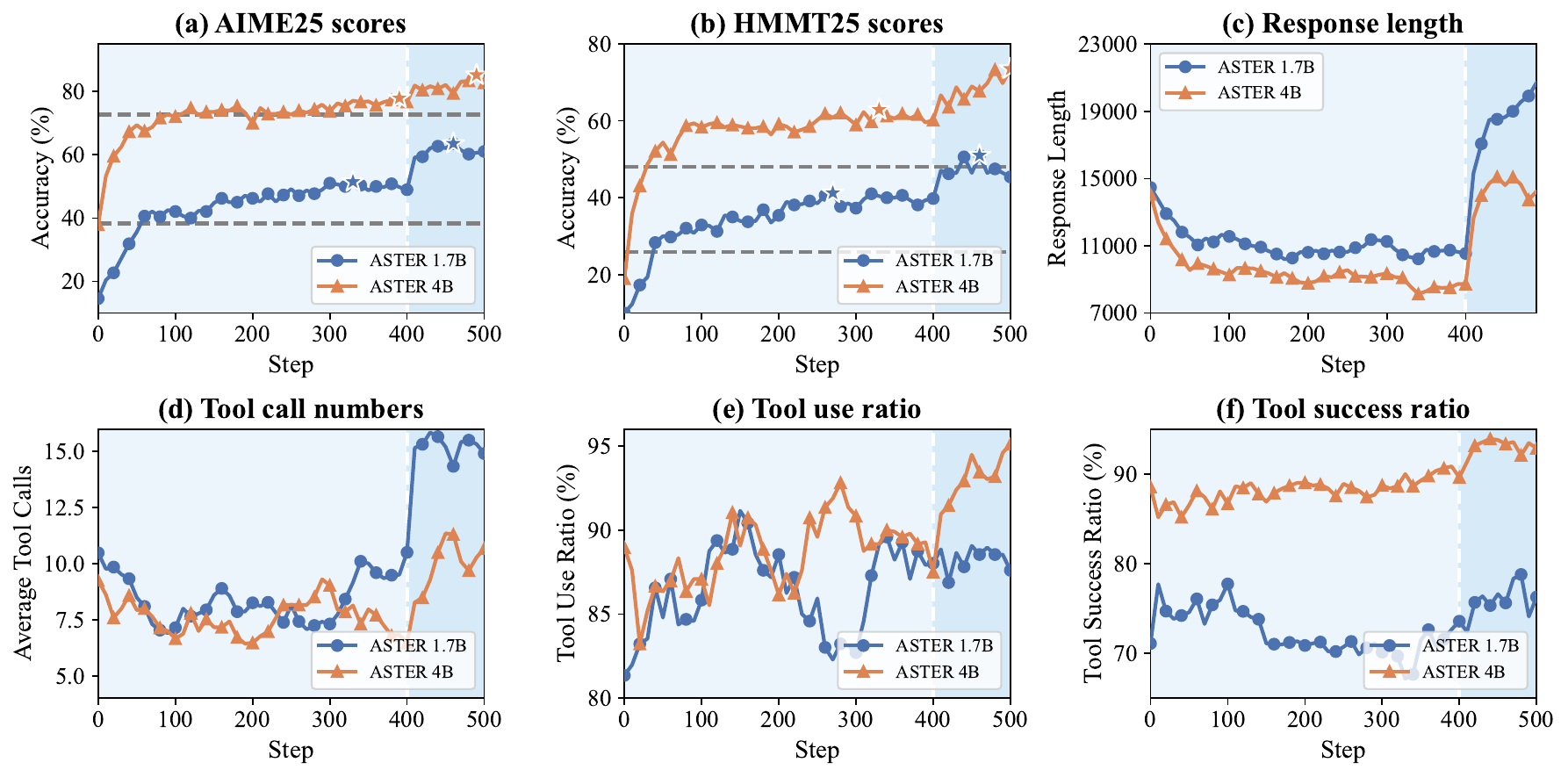}
  \caption{As training progresses, ASTER gradually recovers to pre-SFT performance levels and significantly surpasses them. The model's tool call frequency and response length exhibit an initial decrease followed by an increase, while tool call ratio and execution success rate improve gradually.}  
  \label{fig:main_results}
  \vspace{-15pt}
\end{figure*}

We present comprehensive experimental results across two model scales (1.7B, 4B) in \cref{fig:main_results} to characterize the training dynamics and behavioral patterns of our ASTER framework. Our analysis focuses on three key empirical observations that emerge during the training process.

\paragraph{Cold-start performance recovery} Following cold-start supervised fine-tuning, we observe a marked degradation in model performance on mathematical reasoning benchmarks, with accuracy \textit{substantially falling below} the baseline performance of the pre-SFT base models. 
However, as reinforcement learning training progresses, model accuracy gradually recovers to match and subsequently exceed the pre-SFT baseline levels.

\paragraph{RL training dynamics} As illustrated in \cref{fig:main_results}, during the first RL training stage, we observe a systematic reduction in the average number of tool invocations per trajectory, which stabilizes at a lower plateau after several training epochs. This reduction reflects the model's adaptation to more efficient tool usage patterns, learning to eliminate redundant tool calls while maintaining solution quality. When transitioning to the second training stage with extended context length (32K tokens), we document concurrent improvements across multiple behavioral metrics: average response length increases, tool invocation frequency rises, and benchmark accuracy continues to improve. 
This pattern suggests that the expanded context capacity enables the model to tackle more challenging problems that require longer reasoning trajectories with more frequent tool interactions, while the initial reduction in tool usage establishes a foundation of efficient tool utilization that prevents wasteful invocations.

\paragraph{Model scale efficiency differences} Across different model sizes, we observe systematic differences in tool usage efficiency. Specifically, smaller models (1.7B) exhibit higher tool invocation counts per trajectory compared to medium-scale models (4B). This pattern is mirrored in average response length: smaller models generate longer trajectories, while larger models achieve comparable or superior performance with more concise reasoning paths and fewer tool calls.

\section{Related Works}

TIR enables LLMs to interleave natural-language deliberation with external tool execution (e.g., code interpreters or search engines), improving reliability on tasks requiring precise symbolic or procedural operations.

Prior work often trains TIR via supervised learning on curated tool-use trajectories, e.g., ToRA \cite{gou_tora} and QwenMath-TIR \cite{yang_qwen25math}; however, these SFT pipelines typically rely on high-quality annotations and/or engineered format constraints, which can limit scalability and robustness across tools and domains.

Other studies analyze training dynamics and RL-based tool use: SimpleTIR \cite{xue_simpletir} filters weak-signal trajectories, ZeroTIR \cite{mai_agentrlscalinglaw} characterizes scaling laws for emergent tool invocation, and ARTIST \cite{singh_artist} learns multi-turn tool decisions via outcome-based RL without step-level supervision.

Cold-start synthesis has also been explored: ReTool \cite{feng_retool} rewrites text-only solutions into tool-integrated traces but may break the causal logic of tool invocation, while teacher-generated trajectories \cite{yu_demystifyingreinforcementlearningagentic} can be tool-sparse due to efficient teachers.

In contrast, ASTER seeds outcome-only RL with a small set of \emph{high-interaction} trajectories featuring \emph{causal} plan--execute--interpret--refine loops and explicit \emph{state tracking} in a \emph{stateful} Python sandbox, improving exploration and mitigating collapse to short-horizon verification during multi-stage GRPO training.
\section{Conclusion}

In this work, we introduce ASTER to mitigate \emph{interaction collapse} in tool-integrated reasoning. We show that the behavioral prior from cold-start SFT largely determines RL scaling, and that emphasizing \emph{interaction density}—favoring long-horizon trajectories over immediate post-SFT accuracy—preserves exploration and avoids premature convergence to shallow policies.

Empirical results demonstrate that ASTER-4B achieves state-of-the-art performance, scoring 85.8\% on AIME 2024 and 77.1\% on HMMT 2025. Remarkably, it outperforms significantly larger systems, including Qwen3-235B-A22B-Thinking and rStar2-Agent-14B, proving that compact models can rival giant baselines when equipped with superior tool-integration capabilities.

Our analysis shows that interaction-dense initialization maintains higher training entropy and correlates with final RL performance, highlighting a key principle for agentic scaling: cold-start should prioritize behavioral diversity and exploration over short-term metrics, enabling scalable multi-step problem solving beyond parameter scaling.
\section*{Impact Statement}

This paper studies how to scale tool-integrated reasoning with reinforcement learning and proposes ASTER to mitigate interaction collapse, enabling more reliable multi-step tool use and long-horizon problem solving.
If adopted responsibly, these capabilities may improve the efficiency and accessibility of complex reasoning workflows (e.g., mathematical problem solving and research assistance), potentially reducing reliance on very large models.
At the same time, stronger tool-using agents can amplify misuse risks (e.g., automating harmful tasks) and can introduce safety issues when interacting with external tools or code execution environments.
We therefore emphasize that deployments should use sandboxed tools, least-privilege access, monitoring, and human oversight, and we encourage future work on robust safety evaluation for tool-integrated RL agents.

\bibliography{aster}
\bibliographystyle{icml2026}

\newpage
\appendix
\onecolumn

\section{Experimental Details}

\subsection{Dataset Difficulty Distribution and RL Training Setup}

\begin{figure}[!t]
  \centering
  \begin{subfigure}[b]{0.48\textwidth}
    \centering
    \includegraphics[width=\textwidth]{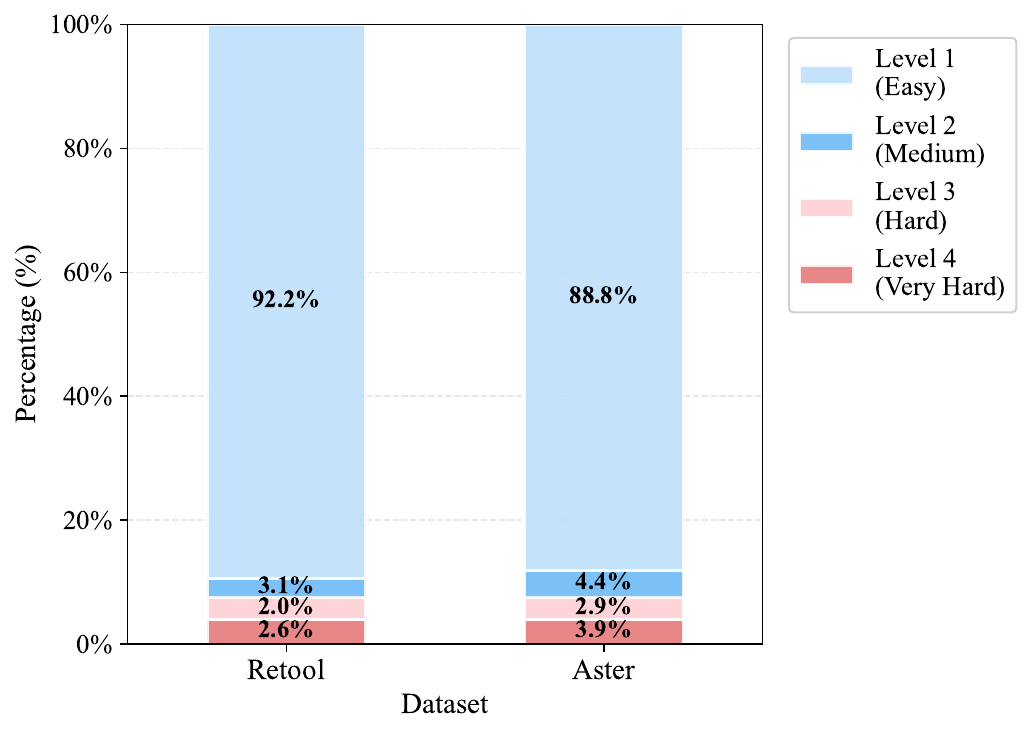}
    \caption{Problem difficulty distribution for the ReTool and ASTER datasets. Difficulty is defined by per-problem accuracy: Easy ($\ge 0.75$), Medium ($[0.50, 0.75)$), Hard ($[0.25, 0.50)$), and Very Hard ($<0.25$).}
    \label{fig:difficulty_distribution_appendix}
  \end{subfigure}
  \hfill
  \begin{subfigure}[b]{0.48\textwidth}
    \centering
    \scalebox{0.88}{
    \begin{tabular}{@{}p{1.8cm}p{4.2cm}p{2.0cm}@{}}
      \toprule
      \bfseries Group & \bfseries Name & \bfseries Value \\
      \midrule
      \multirow{3}{*}{Data}
      & train\_batch\_size & 128 \\
      & max\_prompt\_length & 2048 \\
      & max\_response\_length & 16384/30720 \\
      \addlinespace[0.3em]
      \cmidrule(lr){1-3}
      \addlinespace[0.2em]
      \multirow{6}{*}{Actor}
      & lr & 1e-6 \\
      & ppo\_mini\_batch\_size & 64 \\
      & use\_kl\_loss & False \\
      & clip\_ratio\_low & 0.20 \\
      & clip\_ratio\_high & 0.28 \\
      & entropy\_coeff & 0 \\
      \addlinespace[0.3em]
      \cmidrule(lr){1-3}
      \addlinespace[0.2em]
      \multirow{2}{*}{Rollout}
      & $n$ (group size per prompt) & 8 \\
      & max\_assistant\_turns & 50 \\
      \addlinespace[0.3em]
      \addlinespace[0.3em]
      \bottomrule
    \end{tabular}
    }
    \caption{Hyperparameters used with \texttt{verl}.}
    \label{tab:verl-parameters}
  \end{subfigure}
\end{figure}

To characterize the difficulty distribution of each dataset, we use Qwen3-4B-2507-Thinking as the base model and sample 8 responses per problem. We compute per-problem accuracy as the fraction of correct samples among the 8 responses.
We then stratify problems into four difficulty levels using accuracy thresholds with a step size of 0.25.
As shown in \cref{fig:difficulty_distribution_appendix}, the fraction of easy problems (accuracy in $[0.75, 1.0]$) is 92.2\% and 88.8\% for the two datasets, respectively, suggesting comparable difficulty for the 4B model under the same decoding configuration.

We implement Reinforcement Learning (RL) training using the \texttt{verl} library. Key hyperparameters are summarized in \cref{tab:verl-parameters}.
For the main experiments, we adopt a two-stage training schedule, setting \texttt{max\_response\_length} to 16,384 in Stage~1 and 30,720 in Stage~2. We disable KL regularization to avoid overly constraining long-horizon exploration, especially in the early phase of RL.

In RQ1, for the ablation on different teacher-pattern cold-start strategies, we keep the context length fixed at 32K for both training and evaluation, and hold all other hyperparameters constant.

In RQ2, for the ablation on interaction density, we set the context length to 18K for both RL training and evaluation.

In RQ3, models are trained with an 18K context length, while the maximum number of tool calls is capped at 10 and 50, respectively. During evaluation, we extend the context length to 32K to assess generalization.

\subsection{Additional RQ1 Experimental Results}

To further understand how cold-start strategies shape behavioral priors and affect subsequent RL training, we analyze additional training-time behavioral metrics, including response length and training entropy.

\begin{figure}[htbp]
  \centering
  \begin{subfigure}[b]{0.48\textwidth}
    \centering
    \includegraphics[width=\textwidth]{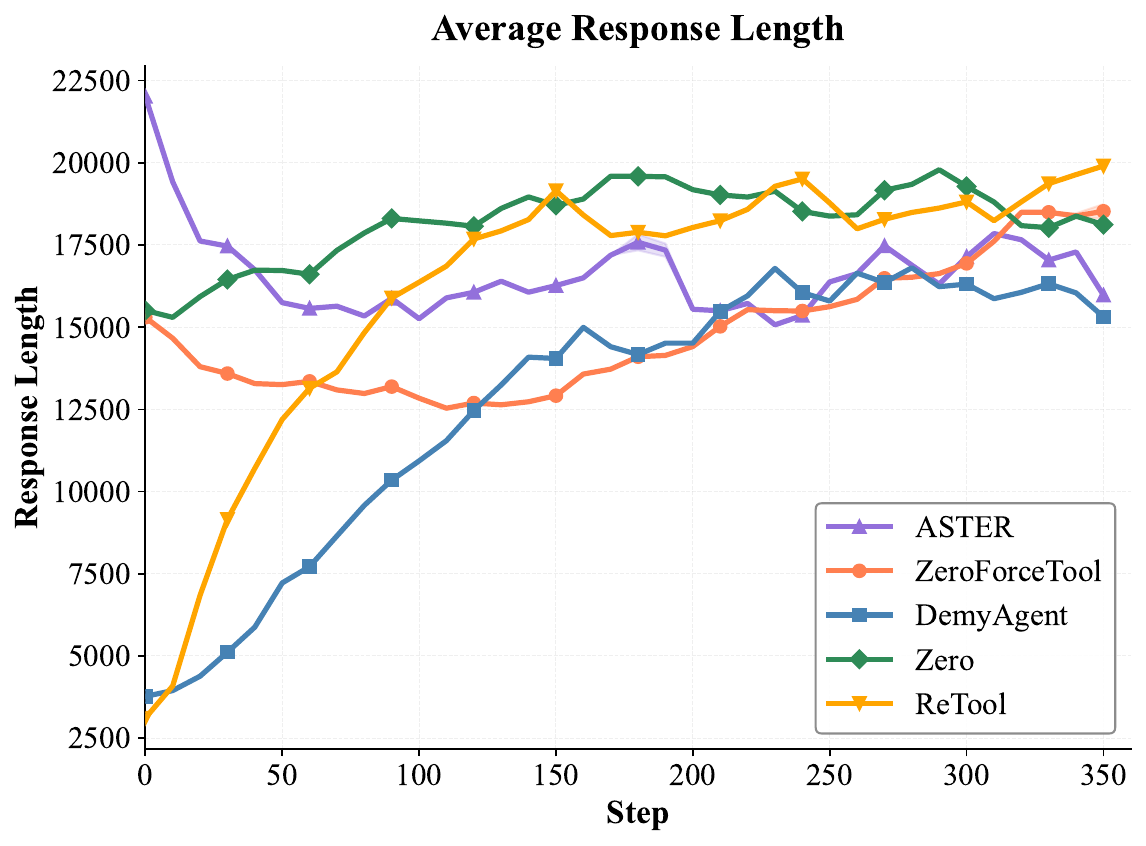}
    \caption{Response length evolution across teacher patterns.}
    \label{fig:appendix_response_length_rq1}
  \end{subfigure}
  \hfill
  \begin{subfigure}[b]{0.48\textwidth}
    \centering
    \includegraphics[width=\textwidth]{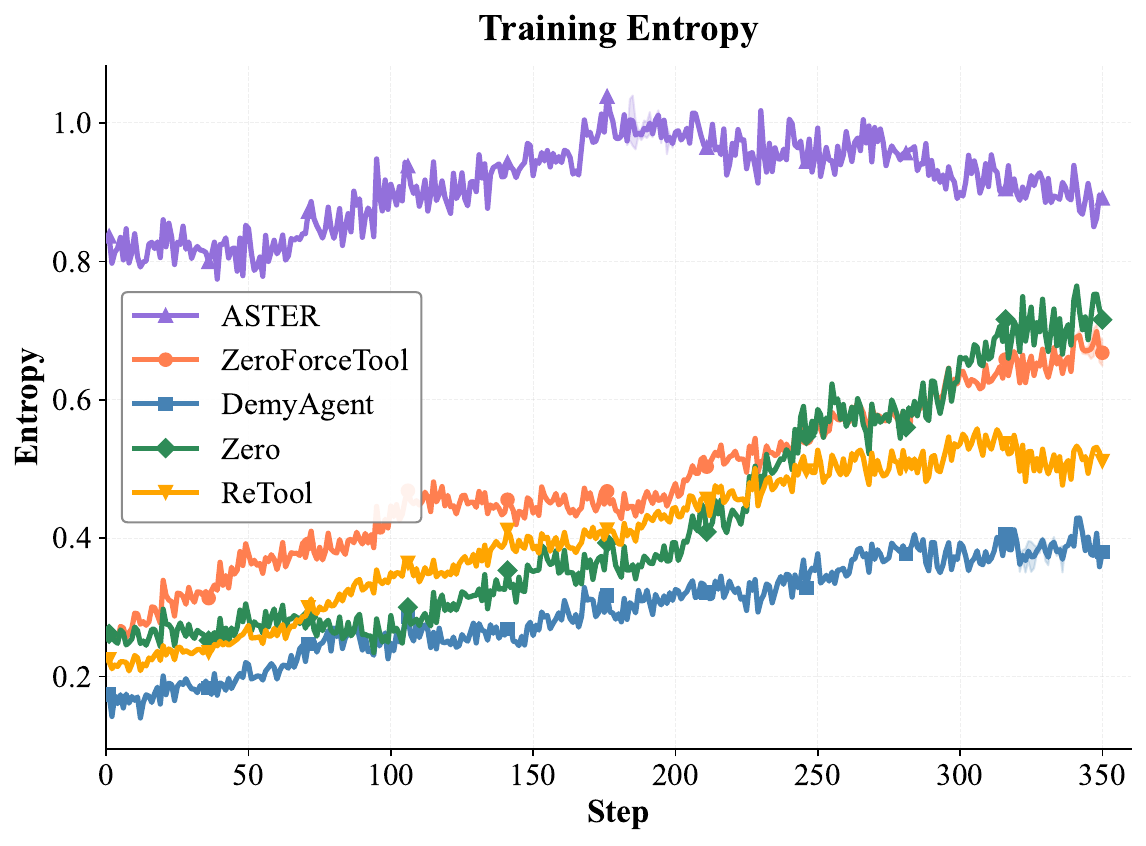}
    \caption{Training entropy evolution across teacher patterns.}
    \label{fig:appendix_training_entropy_rq1}
  \end{subfigure}
  \caption{Training dynamics under different teacher-pattern cold-start strategies. (a) Response length on the validation set. (b) Training entropy over training, reflecting exploration--exploitation dynamics.}
  \label{fig:rq1_appendix_teacher}
  \vspace{-15pt}
\end{figure}

\Cref{fig:rq1_appendix_teacher} shows the training dynamics of models initialized with different teacher-pattern cold starts (Zero, ZeroForceTool, ReTool, DemyAgent, and ASTER), revealing distinct trajectories in response length and training entropy.

Training entropy provides a useful lens on exploration: it captures the uncertainty of the model's generation process during training and serves as a proxy for exploration capacity. Higher entropy indicates broader exploration in the decision space, whereas rapidly declining entropy suggests premature convergence to a narrow behavioral mode.
As shown in \Cref{fig:appendix_training_entropy_rq1}, \textbf{ASTER} sustains the highest training entropy throughout training, suggesting that our interaction-dense cold-start strategy instills a behavioral prior that encourages broader exploration instead of converging early to overly myopic verification-only behaviors. This stronger exploration capacity helps the model better search the solution space during RL, leading to improved final performance.
In contrast, baseline strategies (notably \textbf{Zero} and \textbf{ReTool}) exhibit lower and more rapidly decreasing entropy, consistent with behavioral priors that restrict exploration and induce earlier convergence to suboptimal modes.

\subsection{Additional RQ2 Experimental Results}
\label{sec:appendix_rq2}

\begin{figure}[htbp]
  \centering
  \begin{subfigure}[b]{0.32\textwidth}
    \centering
    \includegraphics[width=\textwidth]{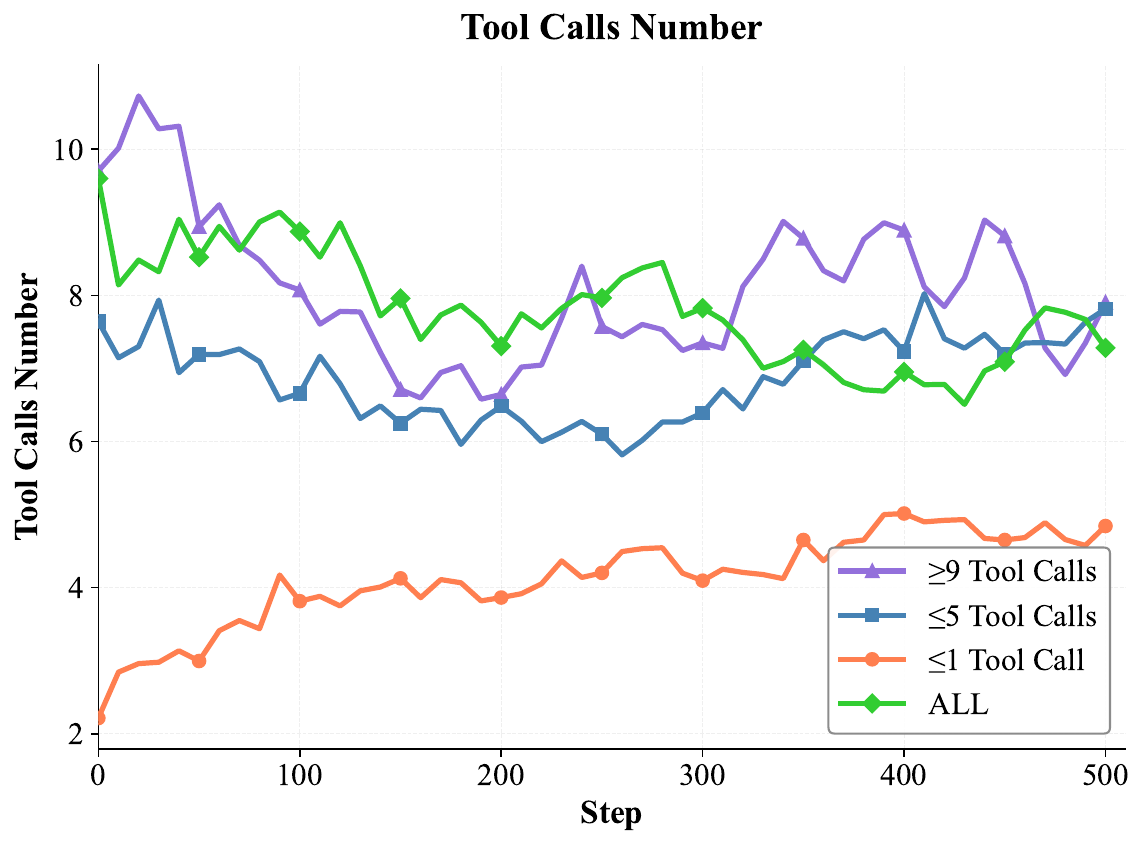}
    \caption{Tool-call count over training.}
    \label{fig:appendix_tool_calls_num_rq2}
  \end{subfigure}
  \hfill
  \begin{subfigure}[b]{0.32\textwidth}
    \centering
    \includegraphics[width=\textwidth]{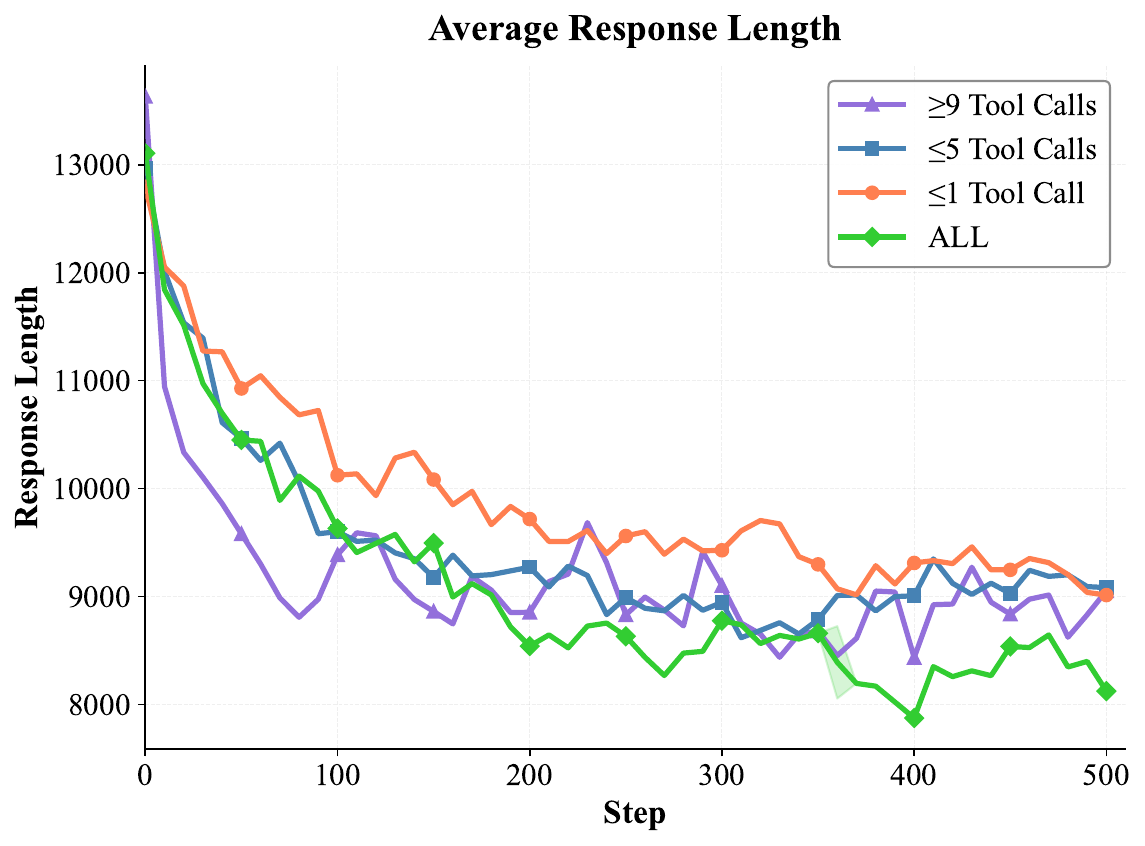}
    \caption{Response length over training.}
    \label{fig:appendix_response_length_rq2}
  \end{subfigure}
  \hfill
  \begin{subfigure}[b]{0.32\textwidth}
    \centering
    \includegraphics[width=\textwidth]{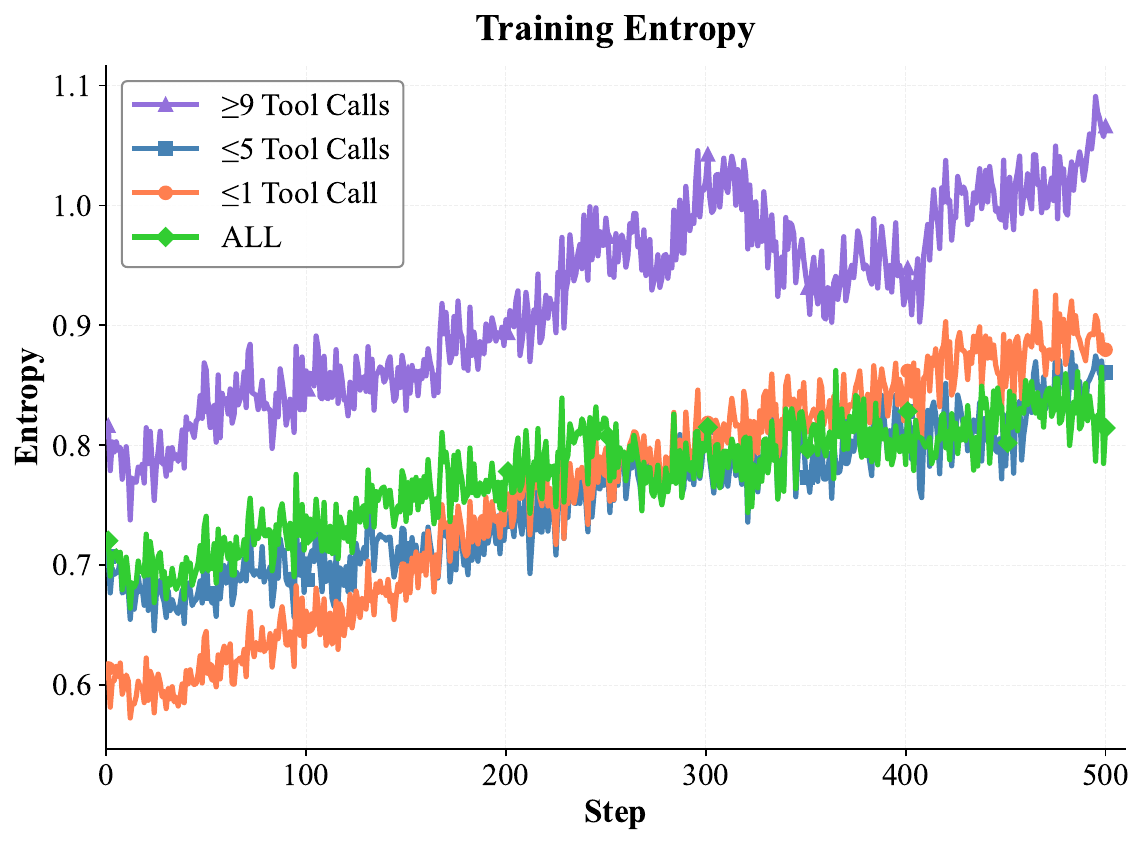}
    \caption{Training entropy over training.}
    \label{fig:appendix_training_entropy_rq2}
  \end{subfigure}
  \caption{Training dynamics under cold starts with different interaction densities. From left to right: tool-call count, response length, and training entropy over training.}
  \label{fig:appendix_density_rq2}
\end{figure}

To further validate the impact of interaction density on behavioral priors, we analyze training dynamics for models cold-started with data stratified by the number of tool calls.
\Cref{fig:appendix_density_rq2} reports the evolution of tool-call counts, response length, and training entropy under different initialization densities.
We find that higher interaction-density cold starts yield higher training entropy, indicating stronger exploration capacity.
Moreover, tool-use patterns during training remain highly consistent with the initial cold-start distribution, demonstrating that interaction density effectively shapes behavioral priors.
Notably, models initialized with high interaction density reduce response length rapidly in the early phase of training, suggesting that tool use can streamline reasoning and improve conciseness.

\subsection{Agentic Judger}\label{sec:agentic_judger}

We use a GPT-5-based agentic judger with a fixed rubric and the following prompt template to score planning, code modeling, error handling, tool efficiency, and overall agentic capability.

\begin{tcolorbox}[
  colback=gray!5,
  colframe=black!60,
  title=\textbf{Prompt Template for Agentic Judger},
  sharp corners,
  boxsep=1.2mm,
  fontupper=\small,
]
You are a strict evaluator for agentic mathematical problem solving. You will be given the COMPLETE TRAJECTORY of an AI Agent, including reasoning, tool calls, code, and intermediate outputs. Your task is to evaluate the Agent's AGENTIC CAPABILITIES, not final answer correctness. Ignore minor syntax errors and formatting issues.

For EACH dimension below:

- Assign an integer score from 1 to 10

- Base your score ONLY on observable behavior in the trajectory

- Use the score anchors as hard references, not vague guidelines

--------------------------------

1. Tool Planning (Score 1--10)

Question: Did the Agent PLAN its tool usage before acting?

Score anchors:

1--2  : No discernible plan. Tool use (if any) is reactive or accidental.  

3--4  : Weak goal awareness. Planning emerges only after failures.  

5--6  : Partial plan exists, but key steps or dependencies are missing.  

7--8  : Explicit multi-step plan. Tool usage is anticipated and sequenced.  

9--10 : Expert-level foresight. Plans include contingencies, checks, and validation steps.

--------------------------------

2. Computational Modeling (Score 1--10)

Question: Did the Agent correctly translate math structure into code?

Score anchors:

1--2  : Uses code as a calculator. No abstraction or structure.  

3--4  : Ad hoc loops or conditionals. Poor alignment with math structure.  

5--6  : Correct brute-force, simulation, or functional decomposition.  

7--8  : Uses appropriate computational paradigms (DP, search, invariants).  

9--10 : Model tightly matches math structure. Efficient, principled, extensible.

--------------------------------

3. Error Diagnosis \& Recovery (Score 1--10)

Question: How well did the Agent detect and fix its own mistakes?

Score anchors:

1--2  : Repeats errors. Ignores signals of failure.  

3--4  : Fixes syntax issues only. Logic errors persist.  

5--6  : Notices incorrect outputs and attempts fixes.  

7--8  : Identifies root causes and tests hypotheses.  

9--10 : Systematic debugging with validation and minimal repro cases.

--------------------------------

4. Tool Use Efficiency (Score 1--10)

Question: Did the Agent use tools when they were MOST appropriate?

Score anchors:

1--2  : Performs long manual reasoning instead of using tools.  

3--4  : Uses tools sporadically; misses obvious opportunities.  

5--6  : Reasonable balance between reasoning and tool use.  

7--8  : Proactive and timely tool usage.  

9--10 : Near-optimal tool choice for all non-trivial operations.

--------------------------------

5. Overall Agentic Quality (Score 1--10)

Question: How strong is the Agent as a TOOL-USING PROBLEM SOLVER?

Guidance:

- This is NOT the average of previous scores.

- Consider coordination between planning, modeling, debugging, and tool use.

Score anchors:

1--3  : Fragmented behavior. Weak planning and execution.  

4--6  : Mixed capability with clear structural weaknesses.  

7--8  : Strong, coherent agentic behavior.  

9--10 : Expert-level autonomy and reasoning-tool synergy.

\end{tcolorbox}

\section{Example Output}

\begin{figure}
    \centering
    \includegraphics[width=1\linewidth]{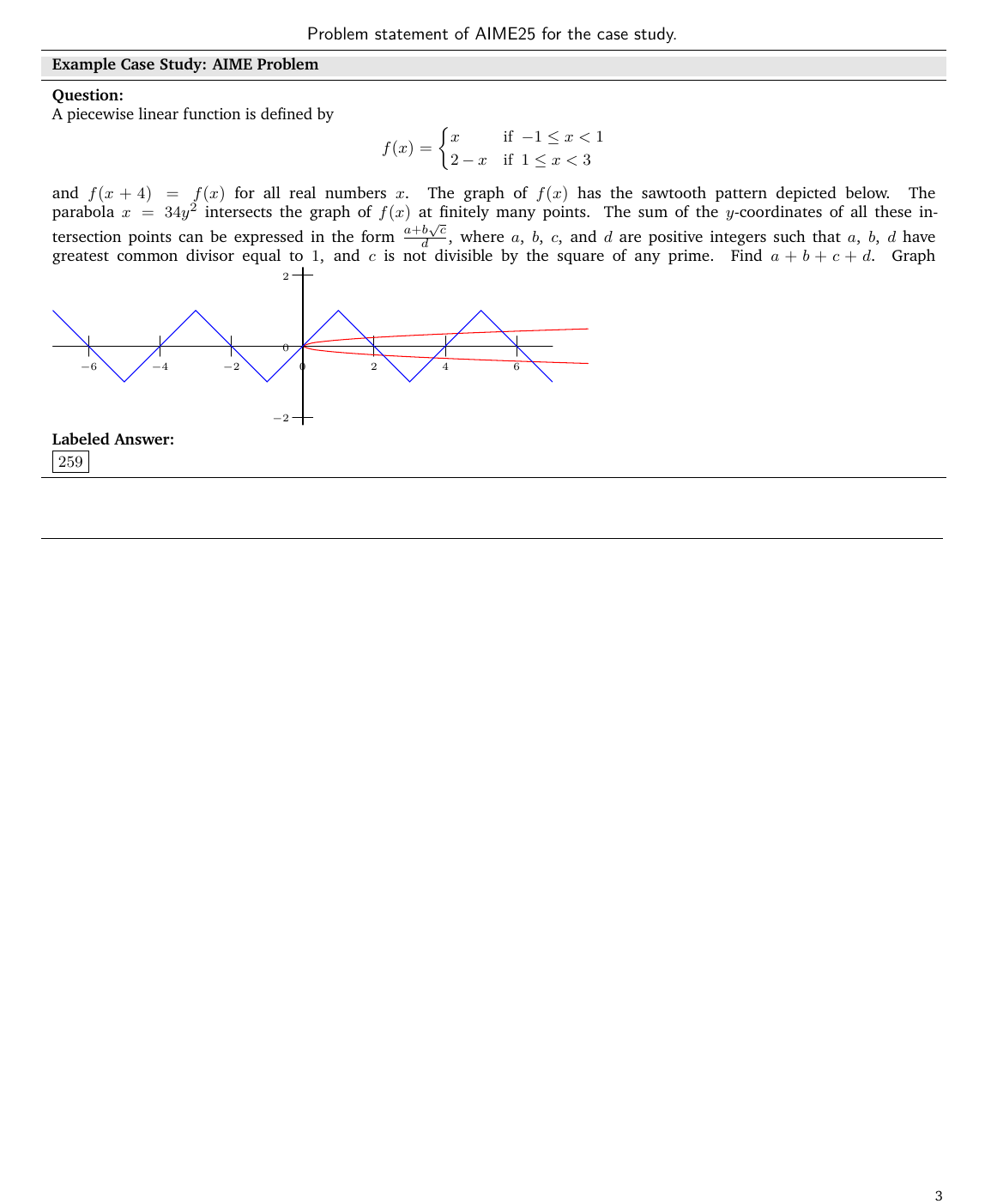}
    \caption{AIME2025-I problem 11}
    \label{fig:demo_1}
\end{figure}

\begin{figure}
    \centering
    \includegraphics[width=1\linewidth]{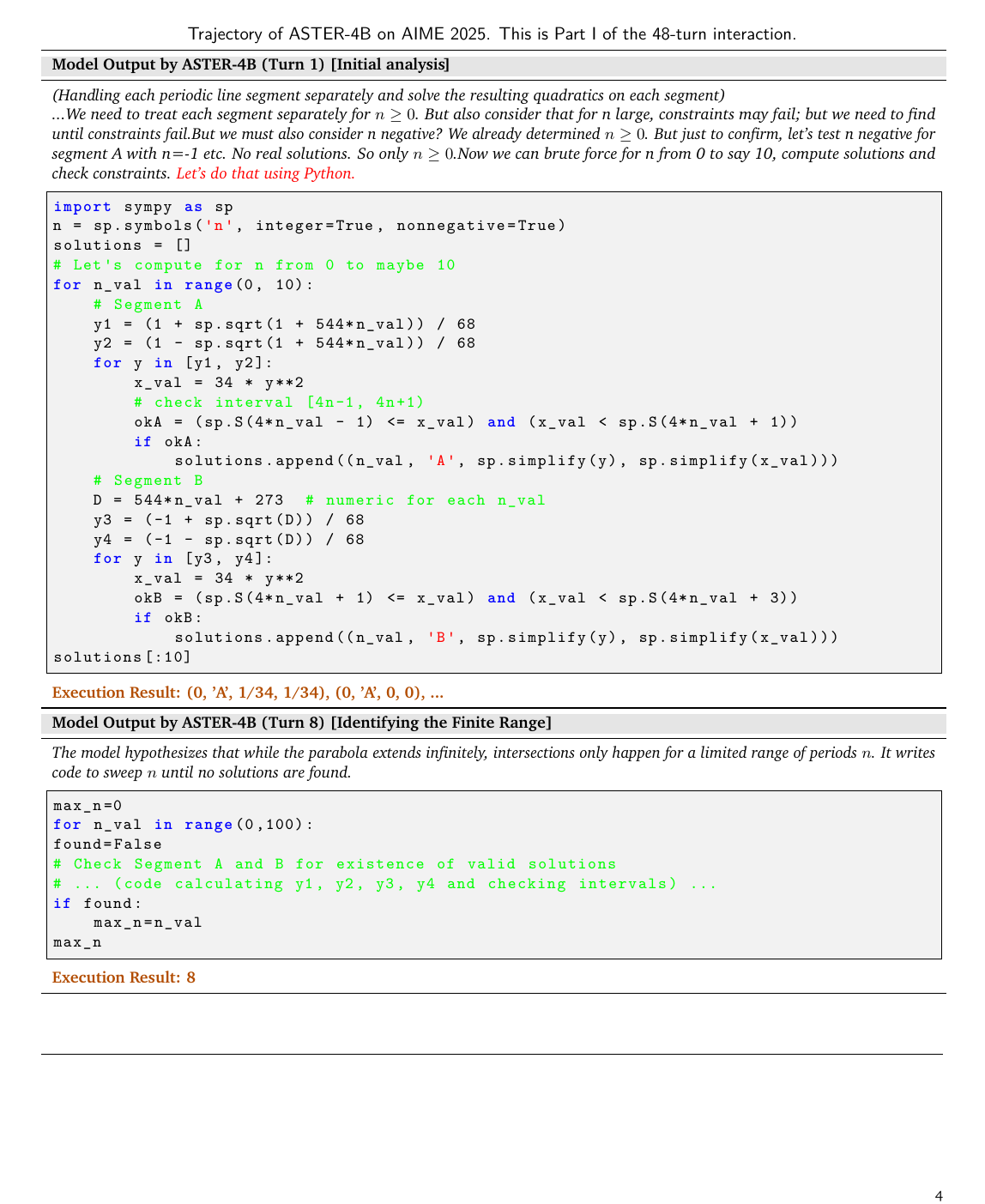}
    \caption*{}
    \label{fig:demo_2}
\end{figure}

\begin{figure}
    \centering
    \includegraphics[width=1\linewidth]{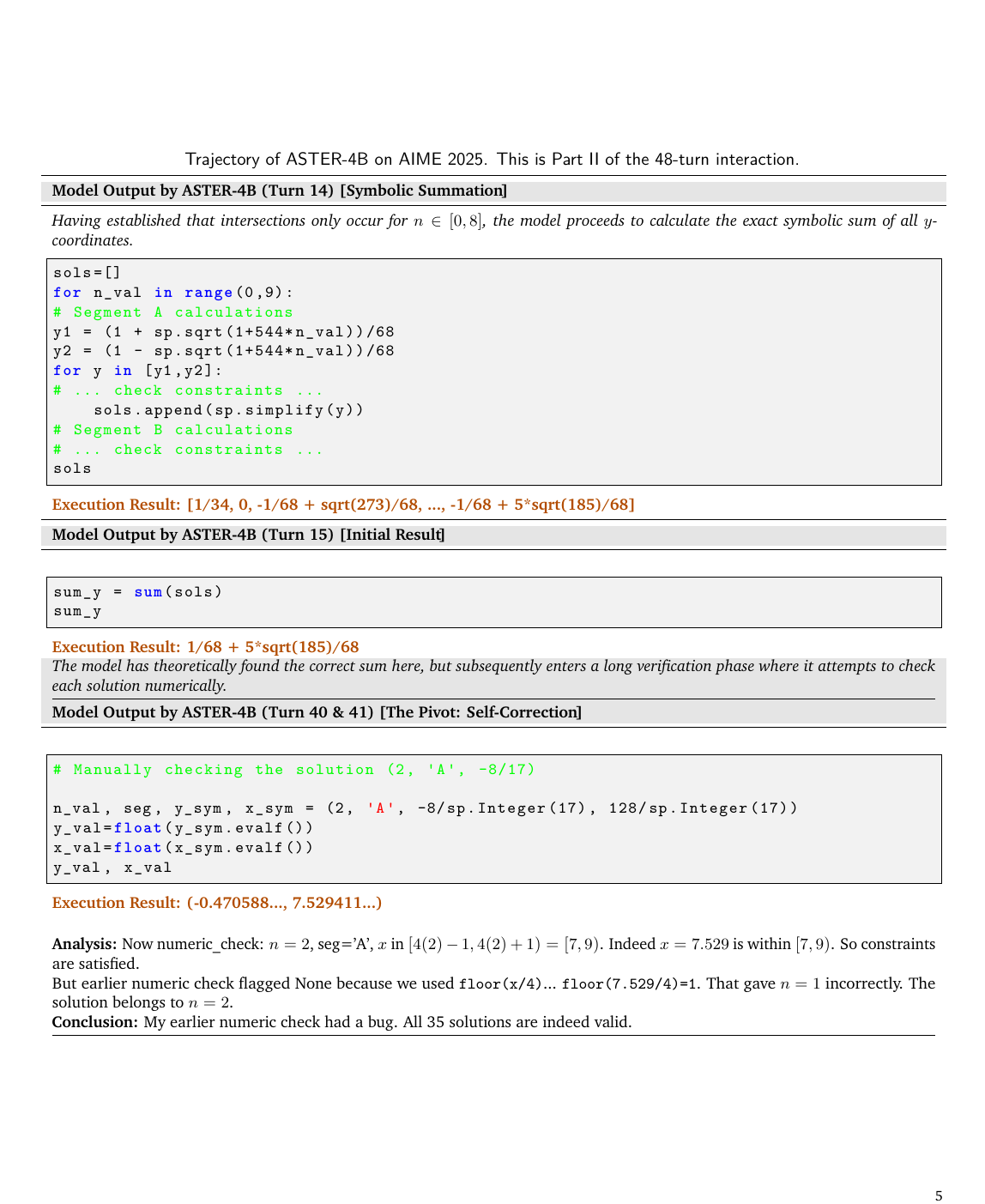}
    \caption*{}
    \label{fig:demo_3}
\end{figure}

\begin{figure}
    \centering
    \includegraphics[width=1\linewidth]{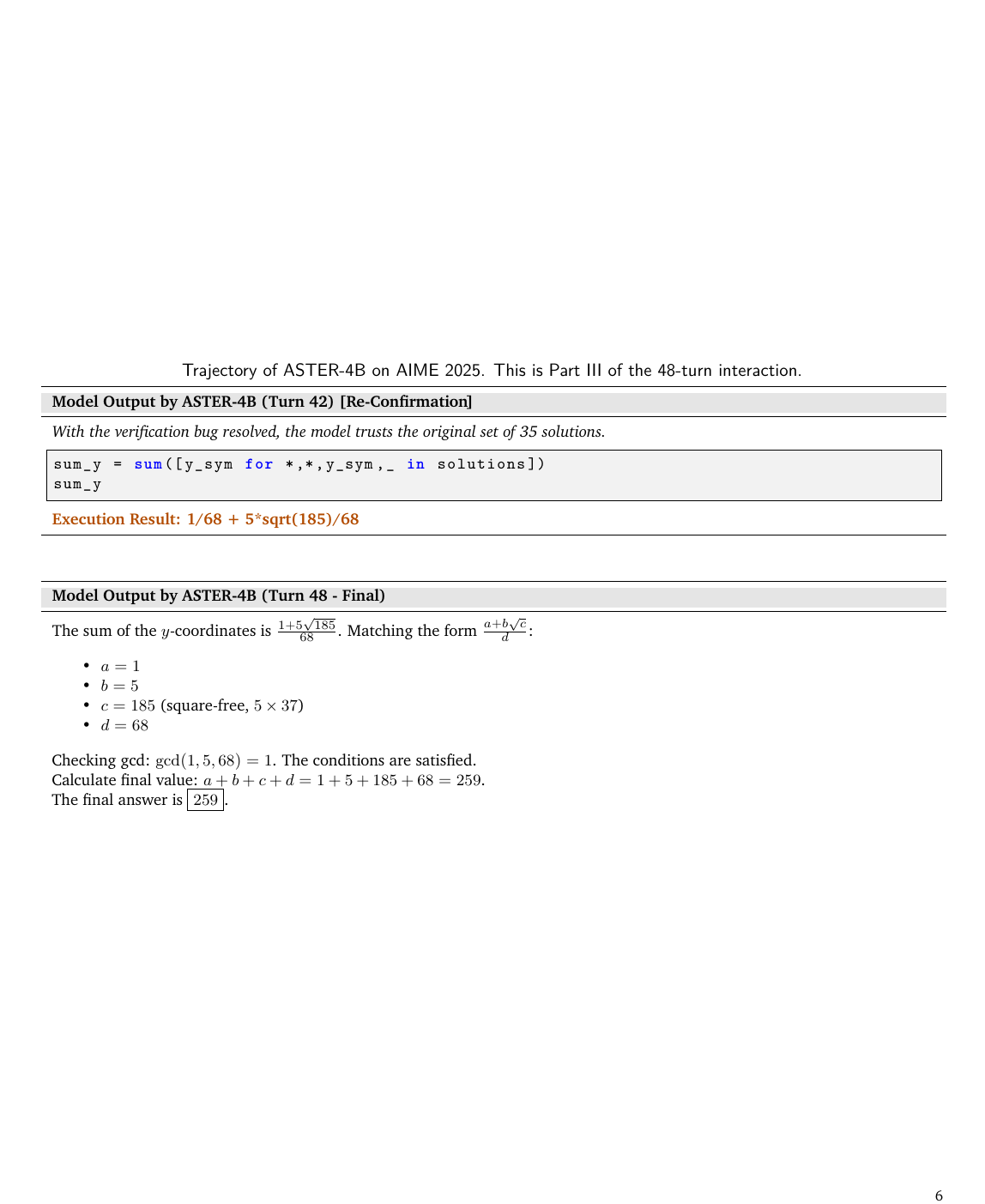}
    \caption*{}
    \label{fig:demo_4}
\end{figure}

\begin{figure}
    \centering
    \includegraphics[width=1\linewidth]{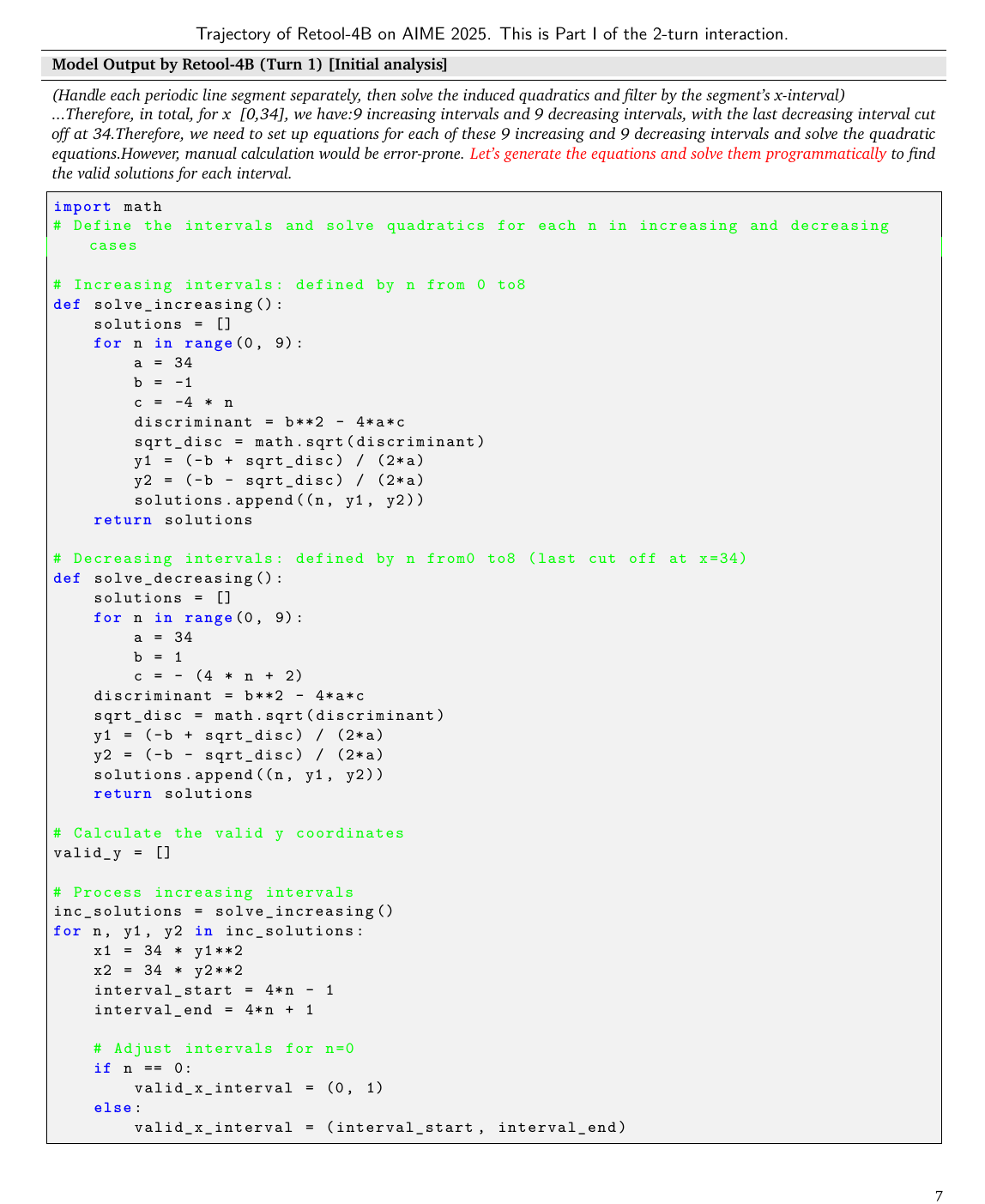}
    \caption*{}
    \label{fig:demo_5}
\end{figure}

\begin{figure}
    \centering
    \includegraphics[width=1\linewidth]{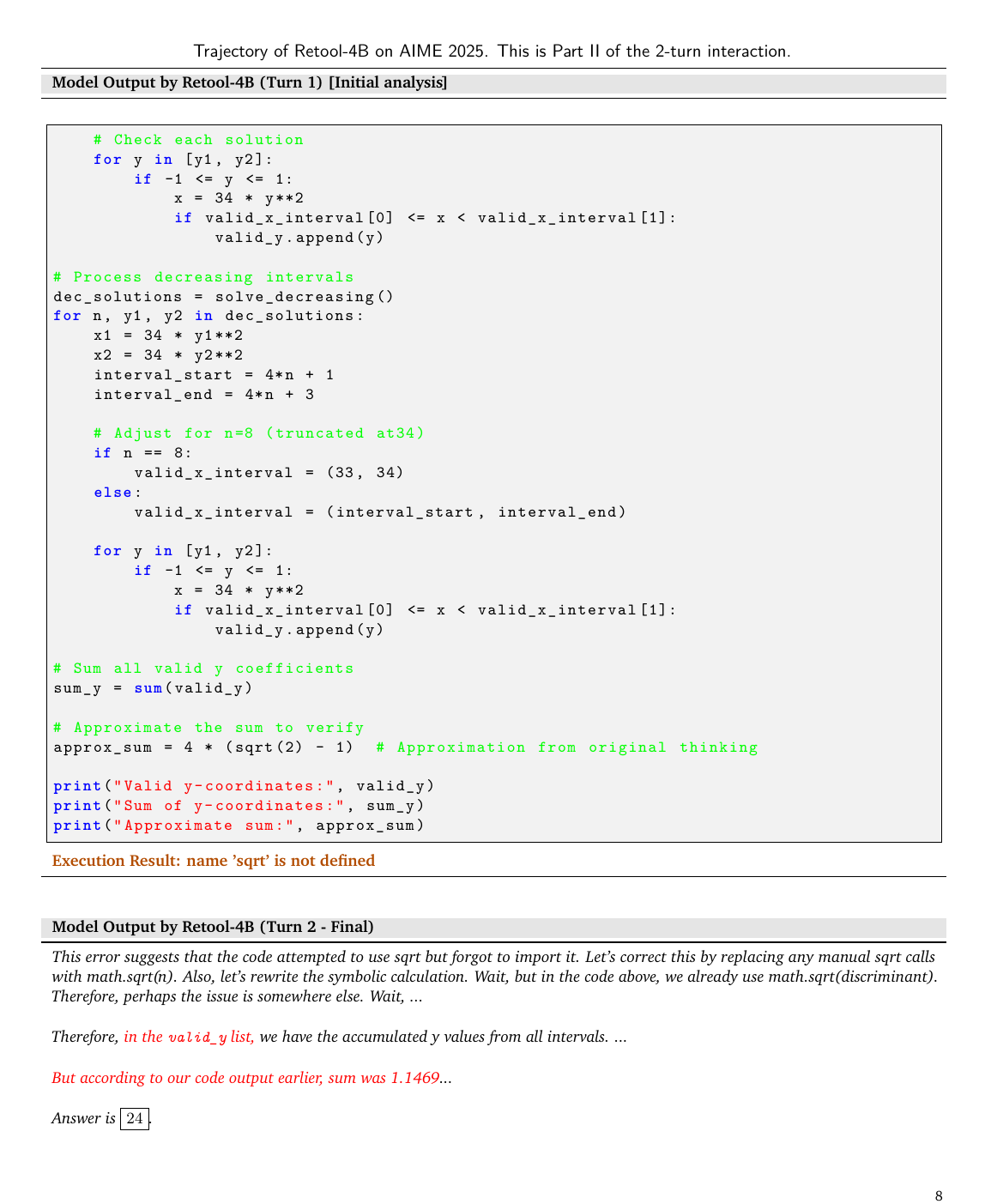}
    \caption*{}
    \label{fig:demo_6}
\end{figure}

\end{document}